\documentclass[runningheads]{llncs}
\usepackage{graphicx}

\usepackage{tikz}
\usepackage{comment}
\usepackage{amsmath,amssymb} 
\usepackage{color}
\usepackage{ctable}
\usepackage{multirow}
\usepackage{pifont}
\usepackage{wrapfig}

\usepackage{comment}
\usepackage{listings}
\usepackage{algorithm}
\usepackage{bbm}
\usepackage{float}
\usepackage[accsupp]{axessibility}  

\usepackage[pagebackref=true,breaklinks=true,colorlinks,bookmarks=false]{hyperref}

\usepackage{breakcites}

\newcommand{\cmark}{\ding{51}}
\newcommand{\xmark}{\ding{55}}
\newcommand{\bigzero}{\mbox{\normalfont\large\bfseries 0}}

\DeclareFontFamily{U}{mathx}{\hyphenchar\font45}
\DeclareFontShape{U}{mathx}{m}{n}{
      <5> <6> <7> <8> <9> <10>
      <10.95> <12> <14.4> <17.28> <20.74> <24.88>
      mathx10
      }{}
\DeclareSymbolFont{mathx}{U}{mathx}{m}{n}
\DeclareFontSubstitution{U}{mathx}{m}{n}
\DeclareMathAccent{\widecheck}{0}{mathx}{"71}

\begin{document}
\pagestyle{headings}
\mainmatter
\def\ECCVSubNumber{0229}  

\title{D\&D: Learning Human Dynamics from\\Dynamic Camera} 


%
\author{Jiefeng Li\inst{1} \and
Siyuan Bian\inst{1} \and
Chao Xu\inst{2} \and
Gang Liu\inst{2} \and
Gang Yu\inst{2} \and
Cewu Lu\inst{1}\thanks{Cewu Lu is the corresponding author, the member of Qing Yuan Research Institute and MoE Key Lab of Artificial Intelligence, AI Institute, Shanghai Jiao Tong University, China and Shanghai Qi Zhi institute}}
\authorrunning{Li et al.}
%
\institute{$^1$~Shanghai Jiao Tong University \quad $^2$~Tencent\\
\email{\{ljf\_likit,biansiyuan,lucewu\}@sjtu.edu.cn}\\
\email{\{dasxu,sylvainliu,skicyyu\}@tencent.com}}
\maketitle

\begin{abstract}
3D human pose estimation from a monocular video has recently seen significant improvements. However, most state-of-the-art methods are kinematics-based, which are prone to physically implausible motions with pronounced artifacts. Current dynamics-based methods can predict physically plausible motion but are restricted to simple scenarios with static camera view. In this work, we present {D\&D} (Learning Human \textbf{D}ynamics from \textbf{D}ynamic Camera), which leverages the laws of physics to reconstruct 3D human motion from the in-the-wild videos with a moving camera. D\&D introduces \textit{inertial force control (IFC)} to explain the 3D human motion in the non-inertial local frame by considering the inertial forces of the dynamic camera. To learn the ground contact with limited annotations, we develop \textit{probabilistic contact torque (PCT)}, which is computed by differentiable sampling from contact probabilities and used to generate motions. The contact state can be weakly supervised by encouraging the model to generate correct motions. Furthermore, we propose an attentive PD controller that adjusts target pose states using temporal information to obtain smooth and accurate pose control. Our approach is entirely neural-based and runs without offline optimization or simulation in physics engines. Experiments on large-scale 3D human motion benchmarks demonstrate the effectiveness of D\&D, where we exhibit superior performance against both state-of-the-art kinematics-based and dynamics-based methods. Code is available at \href{https://github.com/Jeff-sjtu/DnD}{https://github.com/Jeff-sjtu/DnD}.

\end{abstract}

\section{Introduction}

Recovering 3D human pose and shape from a monocular image is a challenging problem. It has a wide range of applications in activity recognition~\cite{li2020detailed,li2020pastanet}, character animation, and human-robot interaction. Despite the recent progress, estimating 3D structure from the 2D observation is still an ill-posed and challenging task due to the inherent ambiguity.

A number of works~\cite{kanazawa2019learning,vibe,luo20203d,choi2021beyond,wan2021encoder,zeng2021smoothnet} turn to temporal input to incorporate body motion priors. Most state-of-the-art methods~\cite{hmr,spin,vibe,moon2020i2l,li2021hybrik,wan2021encoder,yuan2022glamr} are only based on kinematics modeling, i.e., body motion modeling with body part rotations and joint positions. Kinematics modeling directly captures the geometric information of the 3D human body, which is easy to learn by neural networks. However, methods that entirely rely on kinematics information are prone to physical artifacts, such as motion jitter, abnormal root actuation, and implausible body leaning.

Recent works~\cite{rempe2020contact,shimada2020physcap,shimada2021neural,yuan2021simpoe,dabral2021gravity} have started modeling human motion dynamics to improve the physical plausibility of the estimated motion. Dynamics modeling considers physical forces such as contact force and joint torque to control human motion. These physical properties can help analyze the body motion and understand human-scene interaction. Compared to widely adopted kinematics, dynamics gains less attention in 3D human pose estimation. The reason is that there are lots of limitations in current dynamics methods.
For example, existing methods fail in daily scenes with dynamic camera movements (e.g., the 3DPW dataset~\cite{3dpw}) since they require a static camera view, known ground plane and gravity vector for dynamics modeling. Besides, they are hard to deploy for real-time applications due to the need for highly-complex offline optimization or simulation with physics engines.

In this work, we propose a novel framework, D\&D, a 3D human pose estimation approach with learned \textit{Human \textbf{D}ynamics from \textbf{D}ynamic Camera}. Unlike previous methods that build the dynamics equation in the world frame, we re-devise the dynamics equations in the non-inertial camera frame.
Specifically, when the camera is moving, we introduce inertial forces in the dynamics equation to relate physical forces to local pose accelerations. We develop dynamics networks that directly estimate physical properties (forces and contact states). Then we can use the physical properties to compute the pose accelerations and obtain final human motion based on the accelerations.
To train the dynamics network with only a limited amount of contact annotations, we propose \textit{probabilistic contact torque (PCT)} for differentiable contact torque estimation.
Concretely, we use a neural network to predict contact probabilities and conduct differentiable sampling to draw contact states from the predicted probabilities. Then we use the sampled contact states to compute the torque of the ground reaction forces and control the human motion.
In this way, the contact classifier can be weakly supervised by minimizing the difference between the generated human motion and the ground-truth motion.
To further improve the smoothness of the estimated motion, we propose a novel control mechanism called \textit{attentive PD controller}. The output of the conventional PD controller~\cite{shimada2020physcap,shimada2020physcap,yuan2021simpoe} is proportional to the distance of the current pose state from the target state, which is sensitive to the unstable and jittery target. Instead, our attentive PD controller allows accurate control by globally adjusting the target state and is robust to the jittery target.


We benchmark D\&D on the 3DPW~\cite{3dpw} dataset captured with moving cameras and the Human3.6M~\cite{h36m} dataset captured with static cameras. D\&D is compared against both state-of-the-art kinematics-based and dynamics-based methods and obtains state-of-the-art performance.

The contributions of this paper can be summarized as follows:
\begin{itemize}
    \item We present the idea of inertial force control (IFC) to perform dynamics modeling for 3D human pose estimation from a dynamic camera view.
    \item We propose probabilistic contact torque (PCT) that leverages large-scale motion datasets without contact annotations for weakly-supervised training.
    \item Our proposed attentive PD controller enables smooth and accurate character control against jittery target motion.
    \item Our approach outperforms previous state-of-the-art kinematics-based and dynamics-based methods. It is fully differentiable and runs without offline optimization or simulation in physics engines.
\end{itemize}

\section{Related Work}


\subsubsection{Kinematics-based 3D Human Pose Estimation.}
Numerous prior works estimate 3D human poses by locating the 3D joint positions~\cite{akhter2015pose,park20163d,yasin2016dual,moreno20173d,fang2017learning,martinez2017simple,sun2017compositional,pavlakos2017coarse,rogez2017lcr,mehta2017monocular,zhou2017towards,mehta2018single,sun2018integral,moon2019camera,wang2020hmor,zeng2020srnet,li2021localization}.
Although these methods obtain impressive performance, they cannot provide physiological and physical constraints of the human pose.
Many works~\cite{bogo2016keep,lassner2017unite,hmr,spin,moon2020i2l,li2021hybrik} adopt parametric statistical human body models~\cite{loper2015smpl,pavlakos2019expressive,xu2020ghum} to improve physiological plausibility since they provide a well-defined human body structure.
Optimization-based approaches~\cite{bogo2016keep,lassner2017unite,varol2018bodynet,pavlakos2019expressive} automatically fit the SMPL body model to 2D observations, e.g., 2D keypoints and silhouettes.
Alternatively, learning-based approaches use a deep neural network to regress the pose and shape parameters directly~\cite{hmr,spin,vibe,moon2020i2l,li2021hybrik}.
Several works~\cite{lassner2017unite,spin,joo2021exemplar} combine the optimization-based and learning-based methods to produce pseudo supervision or conduct test-time optimization.

For better temporal consistency, recent works have started to exploit temporal context~\cite{mehta2017vnect,kanazawa2019learning,arnab2019exploiting,sun2019human,mehta2020xnect,vibe,choi2021beyond,wan2021encoder,rempe2021humor}.
Kocabas et al.~\cite{vibe} propose an adversarial framework to leverage motion prior from large-scale motion datasets~\cite{mahmood2019amass}. Sun et al.~\cite{sun2019human} model temporal information with a bilinear Transformer. Rempe et al.~\cite{rempe2021humor} propose a VAE model to learn motion prior and optimize ground contacts. All the aforementioned methods disregard human dynamics. Although they achieve high accuracy on pose metrics, e.g., Procrustes Aligned MPJPE, the resulting motions are often physically implausible with pronounced physical artifacts such as improper balance and inaccurate body leaning.



\subsubsection{Dynamics-based 3D Human Pose Estimation.}
To reduce physical artifacts, a number of works leverage the laws of physics to estimate human motion~\cite{yuan2019ego,rempe2020contact,shimada2020physcap,shimada2021neural,yuan2021simpoe,dabral2021gravity}. Some of them are optimization-based approaches~\cite{rempe2020contact,shimada2020physcap,dabral2021gravity}. They use trajectory optimization to obtain the physical forces that induce human motion.
Shimada et al.~\cite{shimada2020physcap} consider a complete human motion dynamics equation for optimization and obtain motion with fewer artifacts. Dabral et al.~\cite{dabral2021gravity} propose a joint 3D human-object optimization framework for human motion capture and object trajectory estimation. Recent works~\cite{zell2020weakly,shimada2021neural} have started to use regression-based methods to estimate human motion dynamics.
Shimada et al.~\cite{shimada2021neural} propose a fully-differentiable framework for 3D human motion capture with physics constraints. All previous approaches require a static camera, restricting their applications in real-world scenarios.

On the other hand, deep reinforcement learning and motion imitation are widely used for 3D human motion estimation~\cite{yuan2019ego,peng2019mcp,yuan2021simpoe,luo2021dynamics,yu2021human}. These works rely on physics engines to learn the control policies. Peng et al.~\cite{peng2019mcp} propose a control policy that allows a simulated character to mimic realistic motion capture data. Yuan et al.~\cite{yuan2021simpoe} present a joint kinematics-dynamics reinforcement learning framework that learns motion policy to reconstruct 3D human motion. Luo et al.~\cite{luo2021dynamics} propose a dynamics-regulated training procedure for egocentric pose estimation. The work of Yu et al.~\cite{yu2021human} is most related to us. They propose a policy learning algorithm with a scene fitting process to reconstruct 3D human motion from a dynamic camera. Training their RL model and the fitting process is time-consuming. It takes $24 \sim 96$ hours to obtain the human motion of one video clip. Unlike previous methods, our regression-based approach is fully differentiable and does not rely on physics engines and offline fitting. It predicts accurate and physically plausible 3D human motion for in-the-wild scenes with dynamic camera movements.


\section{Method}

\begin{figure}[t]
    \begin{center}
    \includegraphics[width=\linewidth]{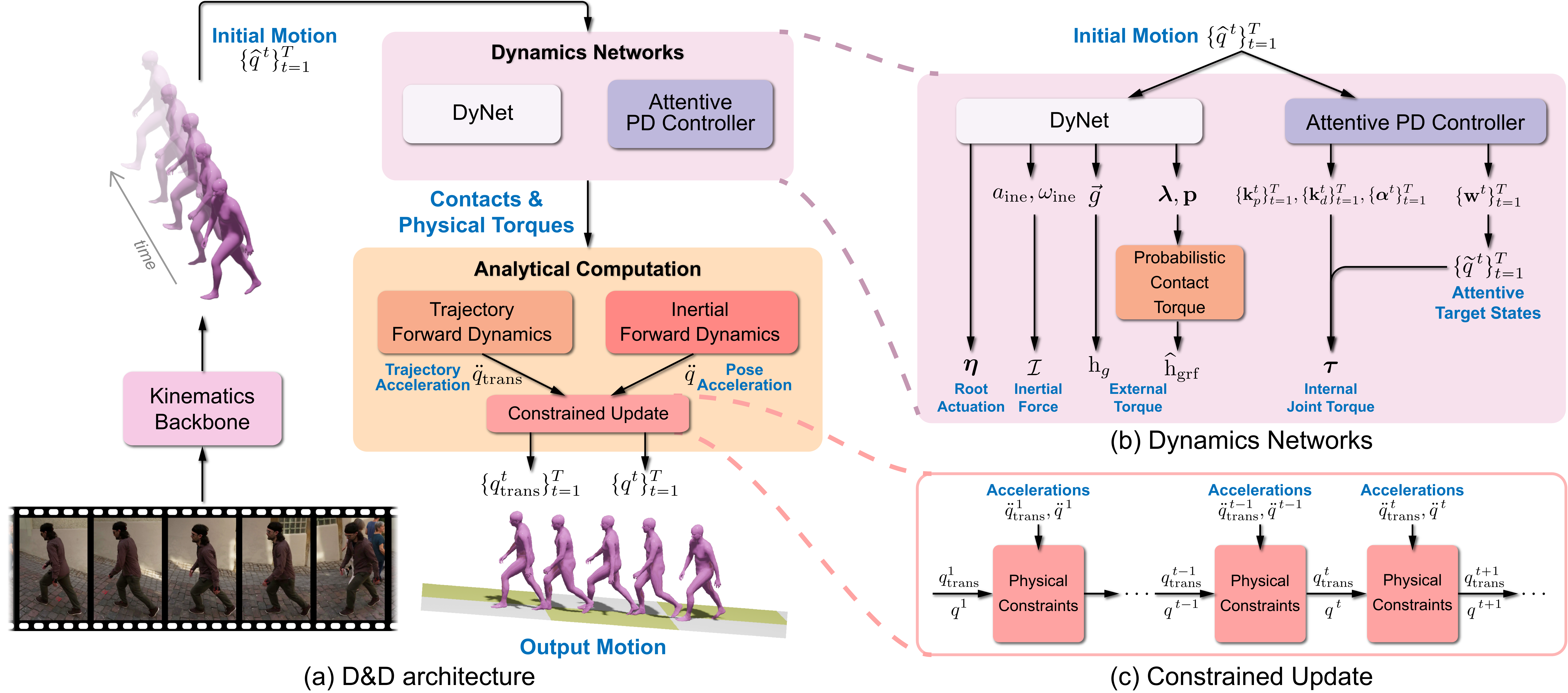}
    \end{center}
    \vspace{-6mm}
    \caption{{\textbf{Overview of the proposed framework.}} The video clip is fed into the kinematics backbone network to estimate the initial motion $\{\widehat{q}^{\,t}\}_{t=1}^{T}$.
    The dynamics networks take as input the initial motion and estimate physical properties. Then we compute pose and trajectory accelerations analytically via forward dynamics. Finally, we utilize accelerations to obtain human motion with physical constraints.
    }
    \label{fig:pipeline}
\end{figure}

The overall framework of the proposed D\&D (\textit{Learning Human \textbf{D}ynamics from \textbf{D}ynamic Camera}) is summarized in Fig.~\ref{fig:pipeline}. The input to D\&D is a video $\{\mathbf{I}^t\}_{t=1}^{T}$ with $T$ frames. Each frame $\mathbf{I}^t$ is fed into the kinematics backbone network to estimate the initial human motion $\widehat{q}^{\,t}$ in the local camera frame. The dynamics networks take as input the initial local motion $\{\widehat{q}^{\,t}\}_{t=1}^{T}$ and estimate physical properties (forces and contact states). Then we apply the forward dynamics modules to compute the pose and trajectory accelerations from the estimated physical properties. Finally, we use accelerations to obtain 3D human motion with physical constraints iteratively.

In this section, before introducing our solution, we first review the formulation of current dynamics-based methods in \S\ref{sec:pre}. In \S\ref{sec:ifc}, we present the formulation of \textit{Inertial Force Control (IFC)} that introduces inertial forces to explain the human motion in the dynamic camera view. Then we elaborate on the pipeline of D\&D: \textbf{i}) learning physical properties with neural networks in \S\ref{sec:learning}, \textbf{ii}) analytically computing accelerations with forward dynamics in \S\ref{sec:forward}, \textbf{iii}) obtaining final pose with the constrained update in \S\ref{sec:update}. The objective function of training the entire framework is further detailed in \S\ref{sec:training}.

\subsection{Preliminaries}
\label{sec:pre}

The kinematic state of the human body can be represented by a pose vector $q$. The pose vector is parameterized as Euler angles with root translation and orientation to build the dynamics formula, i.e., $q \in \mathbb{R}^{3N_j + 6}$, where $N_j$ denotes the total number of human body joints. In previous methods~\cite{shimada2020physcap,yuan2020residual,shimada2021neural}, the first six entries of $q$ are set as the root translation and orientation in the \textit{world frame}. All remaining $3N_j$ entries encode joint angles of the human body. The laws of physics are imposed by considering Newtonian rigid body dynamics as:
\begin{equation}
    \text{M}(q)\ddot{q} - \boldsymbol{\tau} = \text{h}_{\text{grf}}(q, \mathbf{b}, \boldsymbol{\lambda}) - \text{h}_{g}(q, \dot{q}) - \text{h}_{c}(q, \dot{q}),
    \label{eq:rbd1}
\end{equation}
where $\text{M} \in \mathbb{R}^{(3N_j + 6) \times (3N_j + 6)}$ denotes the inertia matrix of the human body; $\dot{q} \in \mathbb{R}^{3N_j + 6}$ and $\ddot{q} \in \mathbb{R}^{3N_j + 6}$ denote the velocity and the acceleration of $q$, respectively;
$\text{h}_{\text{grf}} \in \mathbb{R}^{3N_j + 6}$ denotes the resultant torque of the ground reaction forces;
$\mathbf{b} \in \mathbb{R}^{N_c}$ is the discrete contact states vector;
$\boldsymbol{\lambda} \in \mathbb{R}^{3N_c}$ is the linear contact forces;
$N_c$ denotes the number of joints to which the contact forces are applied;
$\text{h}_g \in \mathbb{R}^{3N_j + 6}$ is the gravity torque; $\text{h}_{c} \in \mathbb{R}^{3N_j + 6}$ encompasses Coriolis and centripetal forces; $\boldsymbol{\tau} \in \mathbb{R}^{3N_j + 6}$ represents the internal joint torque of the human body, with the first six entries being the direct root actuation. In this formulation, the translation and orientation must be in the static world frame, which restricts the application of the model in real-world scenarios. Therefore, previous dynamics-based methods are not applicable in current in-the-wild datasets with moving cameras, e.g., the 3DPW dataset~\cite{3dpw}.

\subsection{Inertial Force Control}
\label{sec:ifc}
In this work, to facilitate in-the-wild 3D human pose estimation with physics constraints, we reformulate the dynamics equation to impose the laws of physics in the dynamic-view video. When the camera is moving, the local frame is an inertial frame of reference. In order to satisfy the force equilibrium, we introduce the inertial force $\mathcal{I}$ in the dynamics system:
\begin{equation}
    \text{M}(q)\ddot{q} - \boldsymbol{\tau} = \text{h}_{\text{grf}}(q, \mathbf{b}, \boldsymbol{\lambda}) - \text{h}_g(q, \dot{q}) - \text{h}_c(q, \dot{q}) + \mathcal{I}(q, \dot{q}, a_{\text{ine}}, \omega_{\text{ine}}),
    \label{eq:rbd2}
\end{equation}
where the first six entries of $q$ are set as the root translation and orientation in the \textit{local camera frame}, and the inertial force $\mathcal{I}$ is determined by the current motion state ($q$ and $\dot{q}$) and camera movement state (linear acceleration $a_{\text{ine}} \in \mathbb{R}^{3}$ and angular velocity $\omega_{\text{ine}} \in \mathbb{R}^{3}$). Specifically, the inertial force encompasses linear, centripetal, and Coriolis forces. It is calculated as follows:
\begin{equation}
    \mathcal{I} = \sum_{i}^{N_j} \underbrace{m_i J_{v_i}^{\mathsf{T}}a_{\text{ine}}}_{\text{linear force} }+ \underbrace{m_i J_{v_i}^{\mathsf{T}}\omega_{\text{ine}}\times(\omega_{\text{ine}} \times r_i)}_{\text{centripetal force}} + \underbrace{2m_i J_{v_i}^{\mathsf{T}} (\omega_{\text{ine}} \times v_i)}_{\text{Coriolis force}},
    \label{eq:inertia}
\end{equation}
where $J_{v_i} \in \mathbb{R}^{3 \times (3N_j + 6)}$ denotes the linear Jacobian matrix that describes how the linear velocity of the $i$-th joint changes with pose velocity $\dot{q}$,
$m_i$ denotes the mass of the $i$-th joint, $r_i$ denotes the position of the $i$-th joint in the local frame, and $v_i$ is the velocity of the $i$-th joint. $J_{v_i}$, $r_i$, and $v_i$ can be analytically computed using the pose $q$ and the velocity $\dot{q}$.




The inertial force control (IFC) establishes the relation between the physical properties and the pose acceleration in the local frame. The pose acceleration can be subsequently used to calculate the final motion. In this way, we can estimate physically plausible human motion from \textit{forces} to \textit{accelerations} to \textit{poses}. The generated motion is smooth and natural. Besides, it provides extra physical information to understand human-scene interaction for high-level activity understanding tasks.

\subsubsection{Discussion.} The concept of \textit{residual force}~\cite{yuan2020residual} is widely adopted in previous works~\cite{levine2012physically,andrews2016real,shimada2020physcap,yuan2020residual,shimada2021neural} to explain the direct root actuation in the global static frame. Theoretically, we can adopt a residual term to explain the inertia in the local camera frame implicitly. However, we found explicit inertia modeling obtains better estimation results than implicit modeling with a residual term. Detailed comparisons are provided in \S\ref{sec:ifrf}.

\subsection{Learning Physical Properties}
\label{sec:learning}

In this subsection, we elaborate on the neural networks for physical properties estimation. We first use a kinematics backbone to extract the initial motion $\{\widehat{q}^{\,t}\}_{t=1}^{T}$. The initial motion is then fed to a dynamics network (DyNet) with \textit{probabilistic contact torque} for contact, external force, and inertial force estimation and the \textit{attentive PD controller} for internal joint torque estimation.



\subsubsection{Contact, External Force, and Inertial Force Estimation.}

The root motion of the human character is dependent on external forces and inertial forces. To explain root motion, we propose DyNet that directly regresses the related physical properties, including the ground reaction forces $\boldsymbol{\lambda} = (\lambda_1, \cdots, \lambda_{N_c})$, the gravity $\vec{g}$, the direct root actuation $\boldsymbol{\eta}$, the contact probabilities $\mathbf{p} = (p_1, p_2, \cdots, p_{N_c})$, the linear camera acceleration $a_{\text{ine}}$, and the angular camera velocity $\omega_{\text{ine}}$. The detailed network structure of DyNet is provided in the supplementary material.

The inertial force $\mathcal{I}$ can be calculated following Eqn.~\ref{eq:inertia} with the estimated $a_{\text{ine}}$ and $\omega_{\text{ine}}$. The gravity torque $\text{h}_g$ can be calculated as:
\begin{equation}
    \text{h}_g = - \sum_i^{N_j} m_i J_{v_i}^{\mathsf{T}} \vec{g}.
\end{equation}
When considering gravity, bodyweight will affect human motion. In this paper, we let the shape parameters $\boldsymbol{\beta}$ control the body weight. We assume the standard weight is $75$kg when $\boldsymbol{\beta}_0 = \mathbf{0}$, and there is a linear correlation between the body weight and the bone length. We obtain the corresponding bodyweight based on the bone-length ratio of $\boldsymbol{\beta}$ to $\boldsymbol{\beta}_0$.

\textit{{Probabilistic Contact Torque}:}
For the resultant torque of the ground reaction forces, previous methods~\cite{shimada2020physcap,yuan2020residual,shimada2021neural} compute it with the discrete contact states $\mathbf{b} = (b_1, b_2, \cdots, b_{N_c})$ of $N_c$ joints:
\begin{equation}
    \text{h}_{\text{grf}}(q, \mathbf{b}, \boldsymbol{\lambda}) = \sum_j^{N_c} b_j J_{v_j}^{\mathsf{T}} \lambda_j,
\end{equation}
where $b_j=1$ for contact and $b_j=0$ for non-contact. Note that the output probabilities $\mathbf{p}$ are continuous. We need to discretize $p_j$ with a threshold of 0.5 to obtain $b_j$. However, the discretization process is not differentiable. Thus the supervision signals for the contact classifier only come from a limited amount of data with contact annotations.


To leverage the large-scale motion dataset without contact annotations, we propose \textit{probabilistic contact torque (PCT)} for weakly-supervised learning. During training, PCT conducts differentiable sampling~\cite{jang2016categorical} to draw a sample $\widehat{\mathbf{b}}$ that follows the predicted contact probabilities $\mathbf{p}$ and computes the corresponding ground reaction torques:
\begin{equation}
    \widehat{\text{h}}_{\text{grf}}(q, \widehat{\mathbf{b}}, \boldsymbol{\lambda}) = \sum_j^{N_c} \widehat{b}_j J_{v_j}^{\mathsf{T}} \lambda_j = \sum_j^{N_c} \frac{p_j e^{g_{j1}}}{p_j e^{g_{j1}} + (1 - p_j)e^{g_{j2}}} J_{v_j}^{\mathsf{T}} \lambda_j,
\end{equation}
where $g_{j1}, g_{j2} \sim \text{Gumbel}(0, 1)$ are i.i.d samples drawn from the Gumbel distribution.
When conducting forward dynamics, we use the sampled torque $\widehat{\text{h}}_{\text{grf}}(q, \widehat{\mathbf{b}}, \boldsymbol{\lambda})$ instead of the torque $\text{h}_{\text{grf}}(q, \mathbf{b}, \boldsymbol{\lambda})$ from the discrete contact states $\mathbf{b}$. To generate accurate motion, DyNet is encouraged to predict higher probabilities for the correct contact states so that PCT can sample the correct states as much as possible.
Since PCT is differentiable, the supervision signals for the physical force and contact can be provided by minimizing the motion error. More details of differentiable sampling are provided in the supplementary material.




\subsubsection{Internal Joint Torque Estimation.}
Another key process to generate human motions is internal joint torque estimation. PD controller is widely adopted for physics-based human motion control~\cite{shimada2020physcap,shimada2021neural,yuan2021simpoe}. It controls the motion by outputting the joint torque $\boldsymbol{\tau}$ in proportion to the difference between the current state and the target state. However, the target pose states estimated by the kinematics backbone are noisy and contain physical artifacts. Previous works~\cite{shimada2021neural,yuan2021simpoe} adjust the gain parameters dynamically for smooth motion control. However, we find that this local adjustment is still challenging for the model and the output motion is still vulnerable to the jittery and incorrect input motion.


\textit{{Attentive PD Controller}:}
To address this problem, we propose the attentive PD controller, a method that allows global adjustment of the target pose states. The attentive PD controller is fed with initial motion $\{\widehat{q}^{\,t}\}_{t=1}^{T}$ and dynamically predicts the proportional parameters $\{\mathbf{k}_p^t\}_{t=1}^{T}$, derivative parameters $\{\mathbf{k}_d^t\}_{t=1}^{T}$, offset torques $\{\boldsymbol{\alpha}^t\}_{t=1}^{T}$, and attention weights $\{\mathbf{w}^t\}_{t=1}^{T}$. The attention weights $\mathbf{w}^t = (w^{t1}, w^{t2}, \cdots, w^{tT})$ denotes how the initial motion contributes to the target pose state at the time step $t$ and $\sum_{j=1}^T w^{tj} = 1$. We first compute the attentive target pose state $\widetilde{q}^{\,t}$ as:
\begin{equation}
    \widetilde{q}^{\,t} = \sum_{j=1}^T w^{tj} \widehat{q}^{\,j},
\end{equation}
where $\widehat{q}^{\,j}$ is the initial kinematic pose at the time step $j$. Then the internal joint torque $\boldsymbol{\tau}^t$ at the time step $t$ can be computed following the PD controller rule with the compensation term $\text{h}_c^t$~\cite{yang2010pd}:
\begin{equation}
    \boldsymbol{\tau}^{t} = \mathbf{k}_p^t \circ (\widetilde{q}^{\,t + 1} - {q}^{t}) - \mathbf{k}_d^t \circ \dot{{q}}^{t} + \boldsymbol{\alpha}^t + \text{h}_c^{t},
\end{equation}
where $\circ$ denotes Hadamard matrix product and $\text{h}_c^t$ represents the sum of centripetal and Coriolis forces at the time step $t$.
This attention mechanism allows the PD controller to leverage the temporal information to refine the target state and obtain a smooth motion. Details of the network structure are provided in the supplementary material.

\subsection{Forward Dynamics}
\label{sec:forward}

To compute the accelerations analytically from physical properties, we build two forward dynamics modules: \textit{inertial forward dynamics} for the local pose acceleration and \textit{trajectory forward dynamics} for the global trajectory acceleration.




\subsubsection{Inertial Forward Dynamics.}
Prior works~\cite{shimada2020physcap,shimada2021neural,yuan2021simpoe} adopt a proxy model to simulate human motion in physics engines or simplify the optimization process. In this work, to seamlessly cooperate with the kinematics-based backbone, we directly build the dynamics equation for the SMPL model~\cite{loper2015smpl}.
The pose acceleration $\ddot{q}$ can be derived by rewriting Eqn.~\ref{eq:rbd2} with PCT:
\begin{equation}
    \ddot{q} = \text{M}^{-1}(q)(\boldsymbol{\tau} + \widehat{\text{h}}_{\text{grf}} - \text{h}_g - \text{h}_c + \mathcal{I}).
\end{equation}
To obtain $\ddot{q}$, we need to compute the inertia matrix $\text{M}$ and other physical torques in each time step using the current pose $q$. The time superscript $t$ is omitted for simplicity.
$\text{M}$ can be computed recursively along the SMPL kinematics tree. The derivation is provided in the supplementary material.






\subsubsection{Trajectory Forward Dynamics.}
To train DyNet without ground-truth force annotations, we leverage a key observation: the gravity and ground reaction forces should explain the global root trajectory. We devise a trajectory forward dynamics module that controls the global root motion with external forces. It plays a central role in the success of weakly supervised learning.

Let $q_{\text{trans}}$ denote the root translation in the \textit{world frame}. The dynamics equation can be written as:
\begin{equation}
    \ddot{q}_{\text{trans}} = \frac{1}{m_0} R^{\mathsf{T}}_{\text{cam}}(\boldsymbol{\eta} + \widehat{\text{h}}_{\text{grf}}^{\{0:3\}} - \text{h}_g^{\{0:3\}}),
\end{equation}
where $m_0$ is the mass of the root joint, $R_{\text{cam}}$ denotes the camera orientation computed from the estimated angular velocity $\omega_{\text{ine}}=(\omega_x, \omega_y, \omega_z)$, $\boldsymbol{\eta}$ denotes the direct root actuation, and $\widehat{\text{h}}_{\text{grf}}^{\{0:3\}}$ and $\text{h}_{\text{g}}^{\{0:3\}}$ denote the first three entries of $\widehat{\text{h}}_{\text{grf}}$ and $\text{h}_{\text{g}}$, respectively.

\subsection{Constrained Update}
\label{sec:update}
After obtaining the pose and trajectory accelerations via forward dynamics modules, we can control the human motion and global trajectory by discrete simulation. Given the frame rate $1/\Delta t$ of the input video, we can obtain the kinematic 3D pose using the finite differences:
\begin{equation}
    \dot{q}^{t + 1} = \dot{q}^t + \Delta t\,\ddot{q}^t,
\end{equation}
\begin{equation}
    q^{t + 1} = q^t + \Delta t\,\dot{q}^t.
    \label{eq:q}
\end{equation}
Similarly, we can obtain the global root trajectory:
\begin{equation}
    \dot{q}_{\text{trans}}^{t + 1} = \dot{q}_{\text{trans}}^t + \Delta t\,\ddot{q}_{\text{trans}}^t,
\end{equation}
\begin{equation}
    q^{t + 1}_{\text{trans}} = q_{\text{trans}}^t + \Delta t\,\dot{q}_{\text{trans}}^t.
    \label{eq:t}
\end{equation}

In practice, since we predict the local and global motions simultaneously, we can impose contact constraints to prevent foot sliding. Therefore, instead of using Eqn.~\ref{eq:q} and \ref{eq:t} to update $q^{t+1}$ and $q^{t + 1}_{\text{trans}}$ directly, we first refine the velocities $\dot{q}^{t + 1}$ and $\dot{q}_{\text{trans}}^{t + 1}$ with contact constraints. For joints in contact with the ground at the time step $t$, we expect they have zero velocity in the world frame. The velocities of non-contact joints should stay close to the original velocities computed from the accelerations. We adopt the differentiable optimization layer following the formulation of Agrawal et al.~\cite{agrawal2019differentiable}. This custom layer can obtain the solution to the optimization problem and supports backward propagation. However, the optimization problem with zero velocity constraints does not satisfy the DPP rules (Disciplined Parametrized Programming), which means that the custom layer cannot be directly applied. Here, we use soft velocity constraints to follow the DPP rules:
\begin{equation}
    \begin{aligned}
        \dot{q}^*, \dot{q}^*_{\text{trans}} = &\mathop{\text{argmin}}_{\dot{q}^*, \dot{q}^*_{\text{trans}}} \| \dot{q}^* - \dot{q} \| + \| \dot{q}^*_{\text{trans}} - \dot{q}_{\text{trans}} \|,\\
        s.t. \quad &\forall i \in \{i | p_i > 0.5\}, ~ \| R_{\text{cam}}^{\mathsf{T}}({J}_{v_i}\dot{q}^* - \dot{q}^{*\{0:3\}}) + \dot{q}^*_{\text{trans}} \| \leq \epsilon,
    \end{aligned}
    \label{eq:optimization}
\end{equation}
where $\epsilon = 0.01$ and $\dot{q}^{*\{0:3\}}$ is the first three entries of $\dot{q}^{*}$. We omit the superscript $t$ for simplicity. After solving Eqn.~\ref{eq:optimization}, the estimated $\dot{q}^*$ and $\dot{q}^*_{\text{trans}}$ are used to compute the final physically-plausible 3D pose $q$ and the global trajectory $q_{\text{trans}}$.

\subsection{Network Training}
\label{sec:training}
The overall loss of D\&D is defined as:
\begin{equation}
    \mathcal{L} =  \mathcal{L}_{\text{3D}} + \mathcal{L}_{\text{2D}} + \mathcal{L}_{\text{con}} + \mathcal{L}_{\text{trans}} + \mathcal{L}_{\text{reg}}.
\end{equation}
The 3D loss $\mathcal{L}_{\text{3D}}$ includes the joint error and the pose error:
\begin{equation}
    \mathcal{L}_{\text{3D}} = \| X - \widecheck{X} \|_1 + \| q \ominus \widecheck{q} \|^2_2,
\end{equation}
where $X$ denotes the 3D joints regressed from the SMPL model, ``$\ominus$'' denotes a difference computation after converting the Euler angle into a rotation matrix, and the symbol ``$\,\widecheck{~}\,$'' denotes the ground truth. The time superscript $t$ is omitted for simplicity. The 2D loss $\mathcal{L}_{\text{2D}}$ calculates the 2D reprojection error:
\begin{equation}
    \mathcal{L}_{\text{2D}} = \| {\rm\Pi}(X) - {\rm\Pi}(\widecheck{X}) \|_1,
\end{equation}
where ${\rm\Pi}$ denotes the projection function. The loss $\mathcal{L}_{\text{trans}}$ is added for the supervision of the root translation and provides weak supervision signals for external force and contact estimation:
\begin{equation}
    \mathcal{L}_{\text{trans}} = \| q_{\text{trans}} - \widecheck{q}_{\text{trans}} \|_1.
\end{equation}
The contact loss is added for the data with contact annotations:
\begin{equation}
    \mathcal{L}_{\text{con}} = \frac{1}{N_c} \sum_i^{N_c} \big[- \widecheck{b}_i \log{p_i} - (1 - \widecheck{b}_i) \log{(1 - p_i})\big].
\end{equation}
The regularization loss $\mathcal{L}_{\text{reg}}$ is defined as:
\begin{equation}
    \mathcal{L}_{\text{reg}} = \| \boldsymbol{\eta} \|^2_2 + \frac{1}{N_c} \sum_i^{N_c} \big[ - p_i \log{p_i} - (1 - p_i) \log{(1 - p_i})\big],
\end{equation}
where the first term minimizes the direct root actuation, and the second term minimizes the entropy of the contact probability to encourage confident contact predictions.

\section{Experiment}

\subsection{Datasets}

We perform experiments on two large-scale human motion datasets. The first dataset is 3DPW~\cite{3dpw}. 3DPW is a challenging outdoor benchmark for 3D human motion estimation. It contains $60$ video sequences obtained from a hand-held moving camera.
The second dataset we use is Human3.6M~\cite{h36m}. Human3.6M is an indoor benchmark for 3D human motion estimation. It includes $7$ subjects, and the videos are captured at 50Hz. Following previous works~\cite{spin,vibe,li2021hybrik,yuan2021simpoe}, we use $5$ subjects (S1, S5, S6, S7, S8) for training and $2$ subjects (S9, S11) for evaluation. The videos are subsampled to 25Hz for both training and testing. We further use the AMASS dataset~\cite{mahmood2019amass} to obtain annotations of foot contact and root translation for training.

\subsection{Implementation Details}
We adopt HybrIK~\cite{li2021hybrik} as the kinematics backbone to provide the initial motion. The original HybrIK network only predicts 2.5D keypoints and requires a separate RootNet~\cite{moon2019camera} to obtain the final 3D pose in the camera frame. Here, for integrity and simplicity, we implement an extended version of HybrIK as our kinematics backbone that can directly predict the 3D pose in the camera frame by estimating the camera parameters. The network structures
are detailed in the supplementary material. The learning rate is set to $5 \times 10^{-5}$ at first and reduced by a factor of $10$ at the $15$th and $25$th epochs. We use the Adam solver and train for $30$ epochs with a mini-batch size of $32$. Implementation is in PyTorch. During training on the Human3.6M dataset, we simulate a moving camera by cropping the input video with bounding boxes.

\begin{table}[t]
    \begin{center}
        \caption{\textbf{Quantitative comparisons with state-of-the-art methods on the 3DPW dataset.} Symbol ``-'' means results are not available, and ``$*$'' means self-implementation.}
        \label{table:3dpw}
        \begin{tabular}{l|ccccc}

        \toprule
        Method & ~Dynamics~ & ~MPJPE~$\downarrow$~ & ~PA-MPJPE~$\downarrow$~ & ~PVE~$\downarrow$~ & ~ACCEL~$\downarrow$~ \\
        \midrule
        HMR~\cite{hmr} & \xmark & 130.0 & 81.3 & - & 37.4 \\
        SPIN~\cite{spin} & \xmark & 96.9 & 59.2 & 116.4 & 29.8 \\
        VIBE~\cite{vibe} & \xmark & 82.9 & 51.9 & 99.1 & 23.4 \\
        TCMR~\cite{choi2021beyond} & \xmark & 86.5 & 52.7 & 102.9 & 7.1 \\
        HybrIK$^*$~\cite{li2021hybrik} & \xmark & 76.2 & 45.1 & 89.1 & 22.8 \\
        MAED~\cite{wan2021encoder} & \xmark & 79.1 & 45.7 & 92.6 & 17.6 \\
        \midrule
        Ours & \cmark & \textbf{73.7} & \textbf{42.7} & \textbf{88.6} & \textbf{7.0} \\

        \bottomrule
        \end{tabular}
    \end{center}
\end{table}

\begin{table}[t]
    \begin{center}
        \caption{\textbf{Quantitative comparisons with state-of-the-art methods on the Human3.6M dataset.} Symbol ``-'' means results are not available, ``$*$'' means self-implementation, and ``$\dagger$'' means the method reports results on $17$ joints.}
        \label{table:h36m}
        \resizebox{\linewidth}{!}
        {
            \begin{tabular}{l|ccccccc}
                \toprule
                Method & ~Dynamics~ & ~MPJPE~$\downarrow$~ & ~PA-MPJPE~$\downarrow$~ & ~PVE~$\downarrow$~ & ~ACCEL~$\downarrow$~ & ~FS~$\downarrow$~ & ~GP~$\downarrow$~ \\
                \midrule
                VIBE~\cite{vibe} & \xmark & 61.3 & 43.1 & - & 15.2 & 15.1 & 12.6 \\
                NeurGD~\cite{song2020human} & \xmark & 57.3 & 42.2 & - & 14.2 & 16.7 & 24.4 \\
                MAED$^\dagger$~\cite{wan2021encoder} & \xmark & 56.3 & 38.7 & - & - & - & - \\
                HybrIK$^*$~\cite{li2021hybrik} & \xmark & 56.4 & 36.7 & - & 10.9 & 18.3 & 10.6  \\
                PhysCap~\cite{shimada2020physcap} & \cmark & 113.0 & 68.9 & - & - & - & - \\
                EgoPose~\cite{yuan2019ego} & \cmark & 130.3 & 79.2 & - & 31.3 & 5.9 & 3.5 \\
                NeurPhys~\cite{shimada2021neural} & \cmark & 76.5 & - & - & - & - & - \\
                SimPoE~\cite{yuan2021simpoe} & \cmark & 56.7 & 41.6 & - & 6.7 & \textbf{3.4} & 1.6 \\
                \midrule
                Ours & \cmark & \textbf{52.5} & \textbf{35.5} & \textbf{72.9} & \textbf{6.1} & 5.8 & \textbf{1.5} \\
                \bottomrule
            \end{tabular}
        }
    \end{center}
\end{table}

\subsection{Comparison to state-of-the-art methods}

\subsubsection{Results on Moving Camera}
We first compare D\&D against state-of-the-art methods on 3DPW, an in-the-wild dataset captured with the hand-held moving camera. Since previous dynamics-based methods are not applicable in the moving camera, prior arts on the 3DPW dataset are all kinematics-based.
Mean per joint position error (MPJPE) and Procrustes-aligned mean per joint position error (PA-MPJPE) are reported to assess the 3D pose accuracy. The acceleration error (ACCEL) is reported to assess the motion smoothness. We also report Per Vertex Error (PVE) to evaluate the entire estimated body mesh.

Tab.~\ref{table:3dpw} summarizes the quantitative results. We can observe that D\&D outperforms the most accurate kinematics-based methods, HybrIK and MAED, by $2.5$ and $5.4$ mm on MPJPE, respectively. Besides, D\&D improves the motion smoothness significantly by $69.3$\% and $60.2$\% relative improvement on ACCEL, respectively. It shows that D\&D retains the benefits of the accurate pose in kinematics modeling and physically plausible motion in dynamics modeling.


\subsubsection{Results on Static Camera} To compare D\&D with previous dynamics-based methods, we evaluate D\&D on the Human3.6M dataset. Following the previous method~\cite{yuan2021simpoe}, we further report two physics-based metrics, foot sliding (FS) and ground penetration (GP), to measure the physical plausibility. To assess the effectiveness of IFC, we simulate a moving camera by cropping the input video with bounding boxes, i.e., the input to D\&D is the video from a moving camera. Tab.~\ref{table:h36m} shows the quantitative comparison against kinematics-based and dynamics-based methods. D\&D outperforms previous kinematics-based and dynamics-based methods in pose accuracy. For physics-based metrics (ACCEL, FS, and GP), D\&D shows comparable performance to previous methods that require physics simulation engines.

We further follow GLAMR~\cite{yuan2022glamr} to evaluate the global MPJPE (G-MPJPE) and global PVE (G-PVE) on the Human3.6M dataset with the simulated moving camera. The root translation is aligned with the GT at the first frame of the video sequence. D\&D obtains 785.1mm G-MPJPE and 793.3mm G-PVE. More comparisons are reported in the supplementary material.

\subsection{Ablation Study}


\begin{figure}[t]
    \begin{center}
    \includegraphics[width=.98\linewidth]{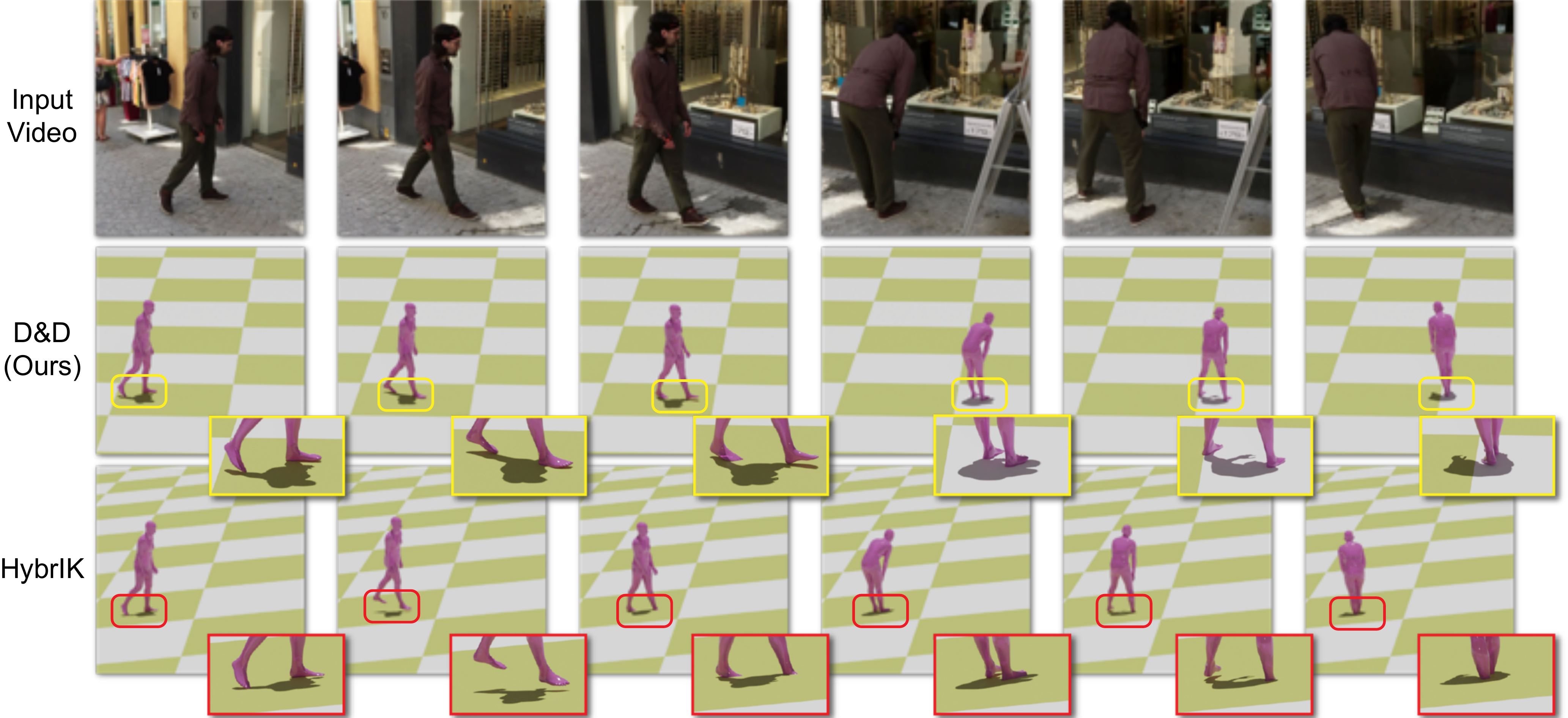}
    \end{center}
    \caption{\textbf{Qualitative comparisons on the 3DPW dataset.} D\&D estimates accurate poses with physically plausible foot contact and global movement.}
    \label{fig:qualitative}
\end{figure}

\begin{table}[t]
    \begin{center}
        \caption{\textbf{Ablation experiments on 3DPW and Human3.6M dataset.}}
        \label{table:ablation}
        \resizebox{\linewidth}{!}
        {
            \begin{tabular}{l|ccc|ccc}
                \toprule
                ~ & \multicolumn{3}{c}{3DPW} & \multicolumn{3}{c}{Human3.6M} \\
                \midrule
                ~ & ~MPJPE~$\downarrow$ & ~PA-MPJPE~$\downarrow$~ & ~ACCEL~$\downarrow$~ & ~MPJPE~$\downarrow$~ & ~PA-MPJPE~$\downarrow$~ & ~ACCEL~$\downarrow$~ \\
                \midrule
                w/o IFC\, & 76.0 & 45.2 & 10.0 & 53.8 & 36.4 & 6.7 \\
                w/o PCT\, & 74.6 & 43.4 & 9.8 & 53.4 & 36.1 & 6.7 \\
                w/o Att PD Controller\, & 73.8 & 42.8 & 8.0 & 52.5 & 35.7 & 6.3 \\
                D\&D (Ours)\, & \textbf{73.7} & \textbf{42.7} & \textbf{7.0} & \textbf{52.5} & \textbf{35.5} & \textbf{6.1} \\
                \bottomrule
            \end{tabular}
        }
    \end{center}
\end{table}

\subsubsection{Inertial Force \textit{vs.} Residual Force.}
\label{sec:ifrf}
In this experiment, we compare the proposed inertial force control (IFC) with residual force control (RFC). To control the human motion with RFC in the local camera frame, we directly estimate the residual force instead of the linear acceleration and angular velocity. Quantitative results are reported in Tab.~\ref{table:ablation}. It shows that explicit modeling of the inertial components can better explain the body movement than implicit modeling with residual force. IFC performs more accurate pose control and significantly reduces the motion jitters, showing a $30$\% relative improvement of ACCEL on 3DPW.

\subsubsection{Effectiveness of PCT.} To study the effectiveness of the probabilistic contact torque, we remove PCT in the baseline model. When training the baseline, the output contact probabilities are discretized to $0$ or $1$ with the threshold of $0.5$ and we compute the discrete contact torque instead of the probabilistic contact torque. Quantitative results in Tab.~\ref{table:ablation} show that PCT is indispensable to have smooth and accurate 3D human motion.

\subsubsection{Effectiveness of Attentive PD Controller.}
To further validate the effectiveness of the attentive mechanism, we report the results of the baseline model without the attentive PD controller. In this baseline, we adopt the meta-PD controller~\cite{yuan2021simpoe,shimada2021neural} that dynamically predicts the gain parameters based on the state of the character, which only allows local adjustment. Tab.~\ref{table:ablation} summarizes the quantitative results. The attentive PD controller contributes to a more smooth motion control as indicated by a smaller acceleration error.



\subsection{Qualitative Results}

\setlength{\intextsep}{0pt}%
\begin{wrapfigure}{r}{0.5\linewidth}
    \centering
    \includegraphics[width=.98\linewidth]{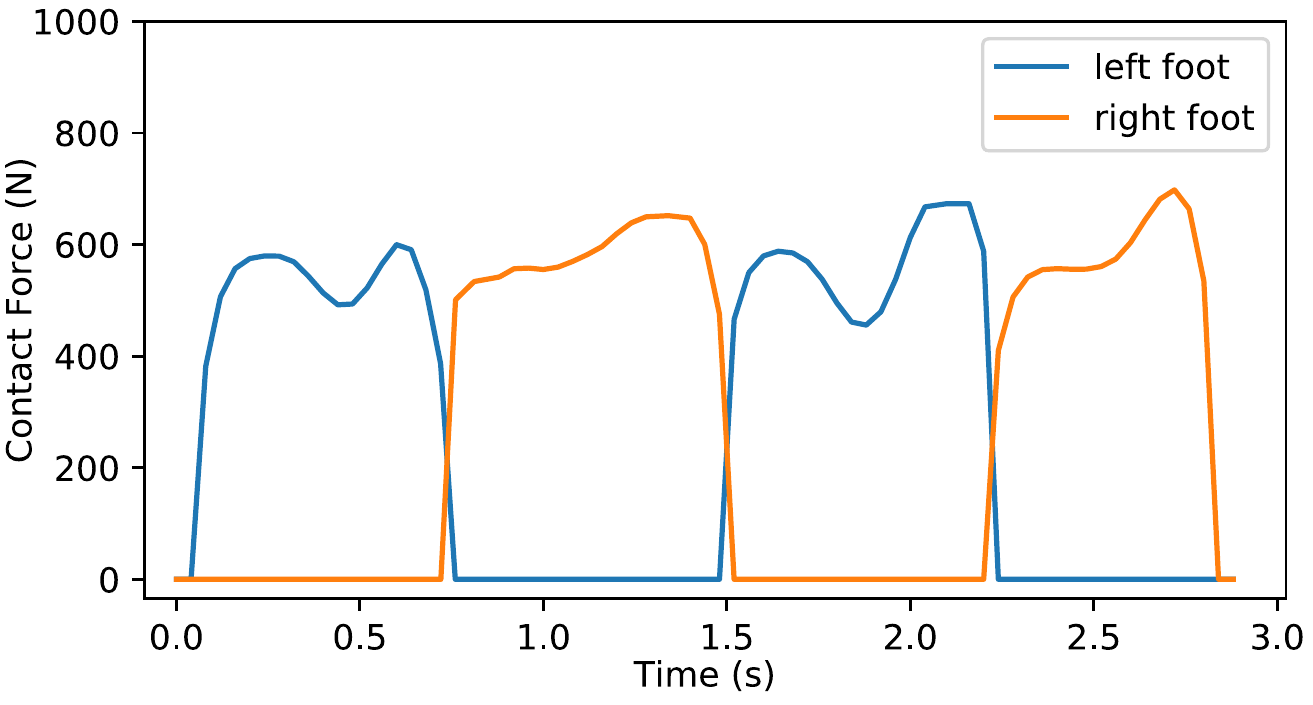}
    \caption{\textbf{Estimated contact forces of the walking sequences.}
    The forces remain in a reasonable range for walking.
    }
    \label{fig:walking}
\end{wrapfigure}

In Fig.~\ref{fig:walking}, we plot the contact forces estimated by D\&D of the walking motion from the Human3.6M test set. Note that our approach does not require any ground-truth force annotations for training. The estimated forces fall into a reasonable force range for walking motions~\cite{shahabpoor2017measurement}. We also provide qualitative comparisons in Fig.~\ref{fig:qualitative}. It shows that D\&D can estimate physically plausible motions with accurate foot-ground contacts and no ground penetration.

\section{Conclusion}
In this paper, we propose D\&D, a physics-aware framework for 3D human motion capture with dynamic camera movements. To impose the laws of physics in the moving camera, we introduce inertial force control that explains the 3D human motion by taking the inertial forces into consideration.
We further develop the probabilistic contact torque for weakly-supervised training and the attentive PD controller for smooth and accurate motion control. We demonstrate the effectiveness of our approach on standard 3D human pose datasets. D\&D outperforms state-of-the-art kinematics-based and dynamics-based methods. Besides, it is entirely neural-based and runs without offline optimization or physics simulators. We hope D\&D can serve as a solid baseline and provide a new perspective for dynamics modeling in 3D human motion capture.

\subsubsection{Acknowledgments.}
This work was supported by the National Key R\&D Program of China (No. 2021ZD0110700), Shanghai Municipal Science and Technology Major Project (2021SHZDZX0102), Shanghai Qi Zhi Institute, SHEITC (2018-RGZN-02046) and Tencent GY-Lab.

\clearpage
%
%
\bibliographystyle{splncs04}
\bibliography{egbib}

\appendix

\section*{Appendix}\label{sec:appendix}

In the supplemental document, we provide:

\begin{itemize}
      \item [\S\ref{sec:derivation}] Derivations of the dynamics quantities.
      \item [\S\ref{sec:sampling}] A more detailed explanation of differentiable sampling.
      \item [\S\ref{sec:arch}] Architectures of the kinematics backbone, DyNet and attentive PD controller.
      \item [\S\ref{sec:contact}] Details for contact annotations.
      \item [\S\ref{sec:exp}] Experiments that compare D\&D with more baselines.
      \item [\S\ref{sec:exp-phys}] Physical-based results on the 3DPW dataset.
      \item [\S\ref{sec:exp-accum}] Global trajectory results on the Human3.6M dataset.
      \item [\S\ref{sec:exp-time}] Computation time of the whole system.
      \item [\S\ref{sec:pseudocode}] \textbf{Pseudocode} of D\&D.
      \item [\S\ref{sec:res}] More qualitative results.
\end{itemize}

\section{Derivations of the Dynamics Quantities}
\label{sec:derivation}

\subsubsection{Inertia Matrix.}

Following Featherstone et al.~\cite{featherstone2014rigid}, the inertia matrix $\text{M}$ is derived as:
\begin{equation}
    \text{M} = \sum_i^{N_j} m_i J_{v_i}^{\mathsf{T}} J_{v_i} + J_{\omega_i}^{\mathsf{T}} I_{c_i} J_{\omega_i},
\end{equation}
where $I_{c_i}$ denotes the inertia tensor of the $i$-th body joint and $J_{\omega_i} \in \mathbb{R}^{3 \times (3N_j + 6)}$ denotes the angular Jacobian matrix that relates angular velocity to pose velocity. $J_{\omega_i}$ can be computed recursively:
\begin{equation}
    J_{\omega_i} = J_{\omega_j} + J_{\omega_{j \rightarrow i}},
\end{equation}
where $j = \mathcal{P}(i)$ is the parent index of the $i$-th joint and $J_{\omega_{j \rightarrow i}}$ represents the relative angular Jacobian matrix that can be computed from $q$.
The linear Jacobian matrix of the $i$-th joint $J_{v_i}$ can be computed based on $J_{\omega_i}$:
\begin{equation}
    J_{v_i} = J_{v_j} - [\Delta r_i]_{\times} J_{\omega_i},
\end{equation}
where $[\Delta r_i]_{\times}$ is the skew-symmetric matrix of the body-part vector $\Delta r_i$. 

\subsubsection{Angular Jacobian Matrix.}

The angular velocity of the $i$-th body joint $\omega_i$ can be computed recursively:
\begin{equation}
    \omega_i = \omega_j + \omega_{j \rightarrow i},
    \label{eq:omega}
\end{equation}
where $j = \mathcal{P}(i)$ is the parent index of the $i$-th joint and $\omega_{j \rightarrow i}$ denotes the relative angular velocity.
Notice that the angular Jacobian matrix of the $i$-th body joint $J_{\omega_i}$ should satisfy:
\begin{equation}
    \omega_i = J_{\omega_i} \dot{q}.
\end{equation}
Therefore, we can build the recursive equation for $J_{\omega_i}$ by replacing $\omega_i$ with $J_{\omega_i}$ in Eqn.~\ref{eq:omega}:
\begin{equation}
    J_{\omega_i} = J_{\omega_j} + J_{\omega_{j \rightarrow i}},
\end{equation}
where
\begin{equation}
    \omega_{j \rightarrow i} = J_{\omega_{j \rightarrow i}} \dot{q}.
\end{equation}
To compute $J_{\omega_i}$, we now need to compute $J_{\omega_{j \rightarrow i}}$ in each step. Denote $(\alpha_i, \beta_i, \gamma_i)$ as the relative rotation of the $i$-th joint in Euler angles.
Then $J_{\omega_{j \rightarrow i}}$ is defined as:
\begin{equation}
    J_{\omega_{j \rightarrow i}} =
        \setlength{\arraycolsep}{3pt}
        \begin{bmatrix}
            \bigzero^{3 \times (3j+3)} & \mathcal{W}_{j \rightarrow i} & \bigzero^{3 \times (3N_j - 3j)} \\
        \end{bmatrix},
\end{equation}
where
\begin{equation}
    \mathcal{W}_{j \rightarrow i} =
    \left[
        \setlength{\arraycolsep}{5pt}
        \renewcommand{\arraystretch}{1.5}
        \begin{array}{ccc}
            \cos{\beta_i}\cos{\gamma_i} & -\sin{\gamma_i} & 0 \\
            \cos{\beta_i}\sin{\gamma_i} & \cos{\gamma_i} & 0 \\
            -\sin{\beta_i} & 0 & 1 \\
        \end{array}
    \right] .
\end{equation}
For the root joint that has no parent, we have:
\begin{equation}
    J_{\omega_0} = \bigzero^{3 \times (3N_j + 6)}.
\end{equation}

\subsubsection{Linear Jacobian Matrix.}
The linear velocity of the $i$-th body joint $v_i$ can be computed based on the angular velocity:
\begin{equation}
    v_i = v_j + \omega_i \times \Delta r_i = v_j - [\Delta r_i]_{\times} \omega_i,
    \label{eq:linear}
\end{equation}
where $\Delta r_i$ denotes the vector of the $i$-th body part. Notice that the linear Jacobian matrix of the $i$-th body joint $J_{v_i}$ should satisfy:
\begin{equation}
    v_i = J_{v_i} \dot{q}.
\end{equation}
Therefore, we can compute $J_{v_i}$ by replacing $v_i$ with $J_{v_i}$ and $\omega_i$ with $J_{\omega_i}$ in Eqn.~\ref{eq:linear}:
\begin{equation}
    J_{v_i} = J_{v_j} - [\Delta r_i]_{\times} J_{\omega_i}.
\end{equation}
For the root joint that has no parent, we have:
\begin{equation}
    J_{v_0} = 
    \setlength{\arraycolsep}{3pt}
    \begin{bmatrix}
        \mathbf{E}^{3 \times 3} & \bigzero^{3 \times (3N_j + 3)} \\
    \end{bmatrix},
\end{equation}
where $\mathbf{E}^{3 \times 3}$ denotes the $3 \times 3$ identity matrix.

\subsubsection{Inertia Tensor.}
Let $I_i$ denote the inertia tensor of the $i$-th body joint under the rest pose that can be pre-computed. The inertia tensor of the $i$-th body joint under the pose $q$ can be computed as:
\begin{equation}
    I_{c_i} = R_i I_i R_i^{\mathsf{T}},
\end{equation}
where $R_i$ denotes the rotation matrix of the $i$-th body joint.


\section{Differentiable Sampling}
\label{sec:sampling}

\subsubsection{Non-differentiable Process.}
The Gumbel-Max trick~\cite{gumbel1954statistical,maddison2014sampling} provides a simple way to draw samples from a categorical distribution. The contact distribution of the $j$-th joint follows the Bernoulli distribution:
\begin{equation}
    \text{Pr}({b}_j = 1) = p_j = 1 - \text{Pr}({b}_j = 0).
\end{equation}
To draw sample $\widehat{b}_j$ with class probability $p_j$, we can conduct:
\begin{equation}
    \widehat{b}_j = \mathop{\arg\max}_k [g_{jk} + \log \text{Pr}(b_j = k)],
\end{equation}
where $k \in \{0, 1\}$. Then the corresponding ground reaction torque can be computed as:
\begin{equation}
    \widehat{\text{h}}_{\text{grf}} = \sum_j^{N_c} \mathbbm{1}(\widehat{b}_j = 1) J_{v_j}^{\mathsf{T}}\lambda_j = \sum_j^{N_c} \widehat{b}_j J_{v_j}^{\mathsf{T}}\lambda_j.
\end{equation}

\subsubsection{Differentiable Process.}
However, Gumbal-Max is not differentiable. Therefore, we adopt Gumbel-Softmax~\cite{jang2016categorical} to conduct differentiable sampling from the predicted distribution. Gumbel-Softmax is a continuous and differentiable approximation to Gumbel-Max by replacing \verb+argmax+ with the softmax function:
\begin{equation}
    \widehat{b}_j = \frac{ \exp(\log \text{Pr}(b_j = 1) + g_{j1} )}{ \sum_{k \in \{0, 1\}} \exp(\log \text{Pr}(b_j = k) + g_{jk} )}.
\end{equation}
Therefore, the corresponding ground reaction torque can be computed as:
\begin{equation}
    \widehat{\text{h}}_{\text{grf}} = \sum_j^{N_c} \frac{p_j e^{g_{j1}}}{p_j e^{g_{j1}} + (1 - p_j)e^{g_{j0}}} J_{v_j}^{\mathsf{T}} \lambda_j,
\end{equation}

\begin{figure}[t]
    \centering
    \includegraphics[width=.98\linewidth]{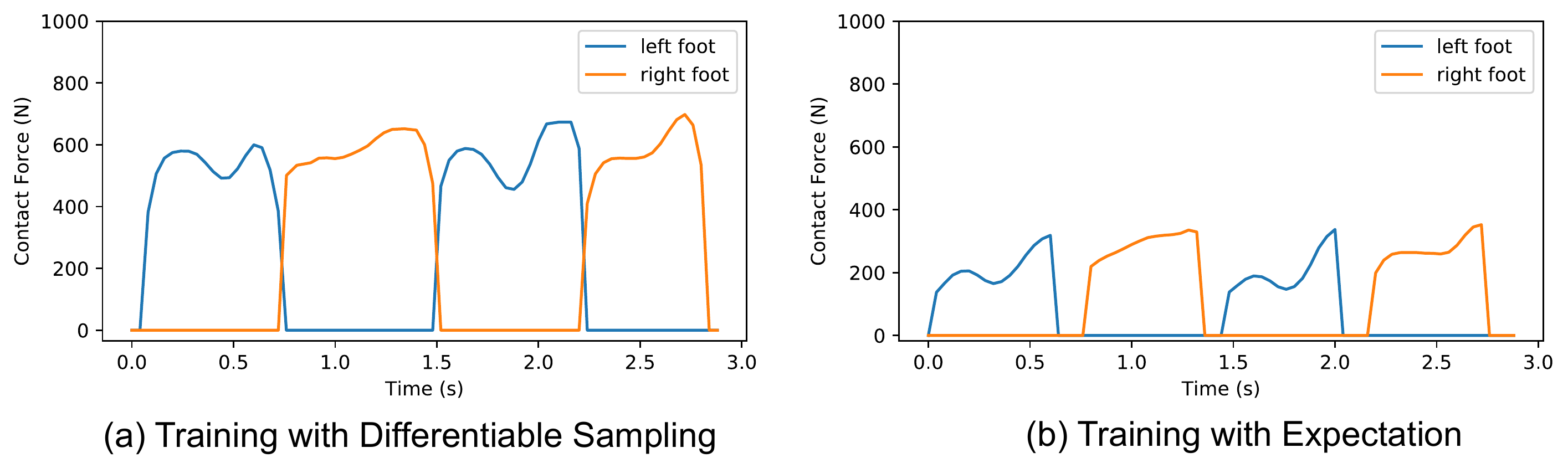}
    \caption{\textbf{Estimated contact forces of the walking sequence.}
    }
    \label{fig:force}
\end{figure}

Alternatively, we can compute the expectation of the ground reaction torque, which is also differentiable:
\begin{equation}
    \bar{\text{h}}_{\text{grf}} = \sum_j^{N_c} p_j J_{v_j} \lambda_j.
\end{equation}
However, using the expectation to generate motion cannot encourage well-calibrated probabilities~\cite{li2021localization}, i.e., DyNet is not encouraged to generate high probabilities for the correct contact states. Consequently, the contact forces might be incorrect since the model is trained without direct supervision. We plot the contact forces of the walking motion using the model trained with the torque expectation in Fig.~\ref{fig:force}(b). It shows that using the torque expectation makes the contact forces lie in an unreasonable range.

\section{Network Architecture}
\label{sec:arch}

\subsection{Kinematics Backbone}

\begin{figure}[h]
    \begin{center}
    \includegraphics[width=.8\linewidth]{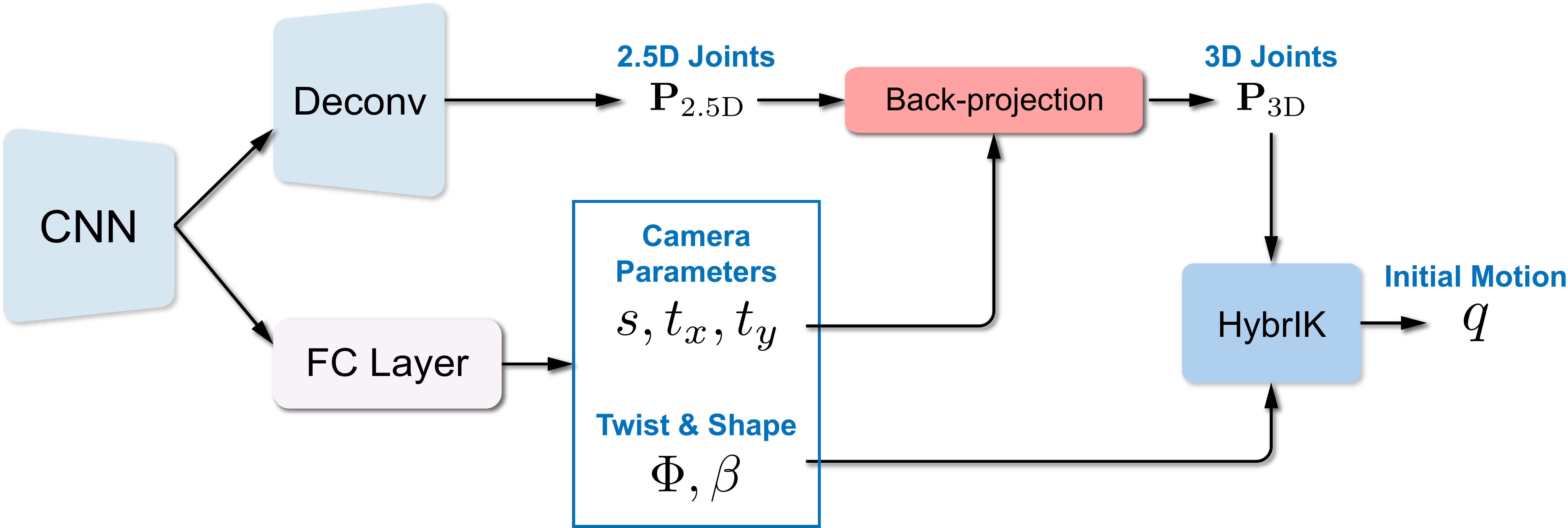}
    \end{center}
    \caption{\textbf{The detailed architecture of the kinematics backbone.}}
    \label{fig:hybrik}
\end{figure}

The detailed network architecture of the kinematics backbone is illustrated in Fig.~\ref{fig:hybrik}. We implement an extended version of HybrIK~\cite{li2021hybrik} as the backbone network. The original HybrIK model first predicts 2.5D joints of the body skeleton. Then the RootNet~\cite{moon2019camera} is adopted to predict the distance of the root joint to the camera plane and obtain the 3D joints in the camera frame via back-projection. In our implementation, we design an integrated model that directly estimates the 3D joint positions. Specifically, we add a fully-connected layer to regress the camera parameters $(s, t_x, t_y)$. Therefore, we can obtain the 3D joint positions as well as the initial motion within a single model.

\subsection{DyNet}

\begin{figure}[h]
    \begin{center}
    \includegraphics[width=.9\linewidth]{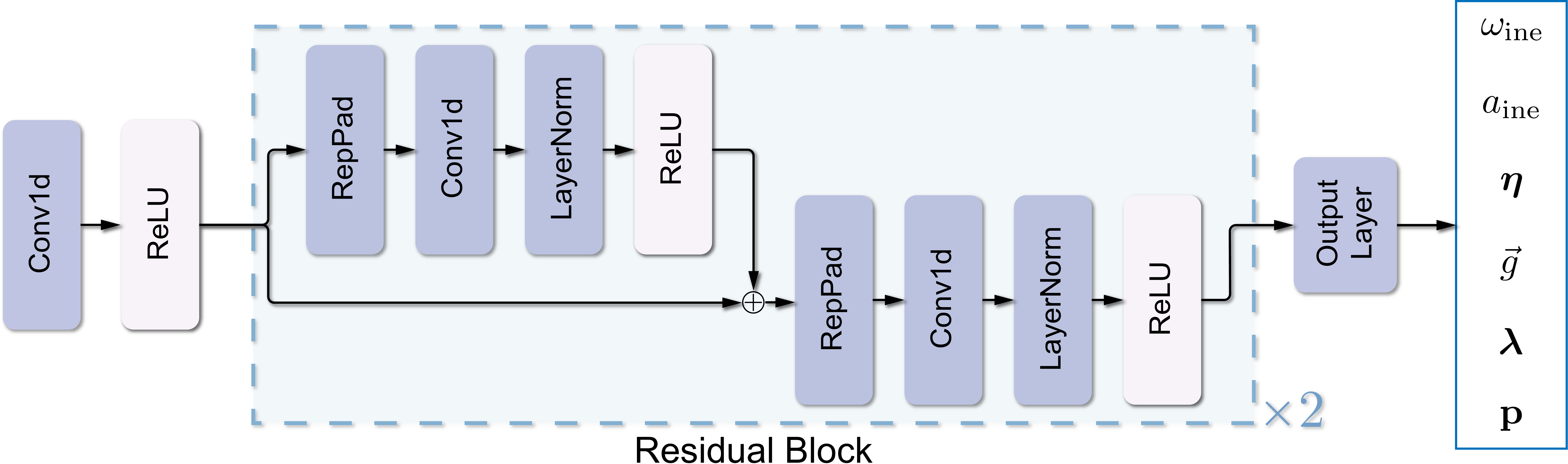}
    \end{center}
    \caption{\textbf{The detailed architecture of DyNet.}}
    \label{fig:dynet}
\end{figure}

The detailed network architecture of the DyNet is illustrated in Fig.~\ref{fig:dynet}. For all 1D convolutional layers, we use the kernel of size $3$ and channel of size $256$. The output layer consists of a bilinear GRU layer with the hidden dimension of $1024$ and a fully-connected layer to predict the physical properties.

\subsection{Attentive PD Controller}

\begin{figure}[h]
    \begin{center}
    \includegraphics[width=.9\linewidth]{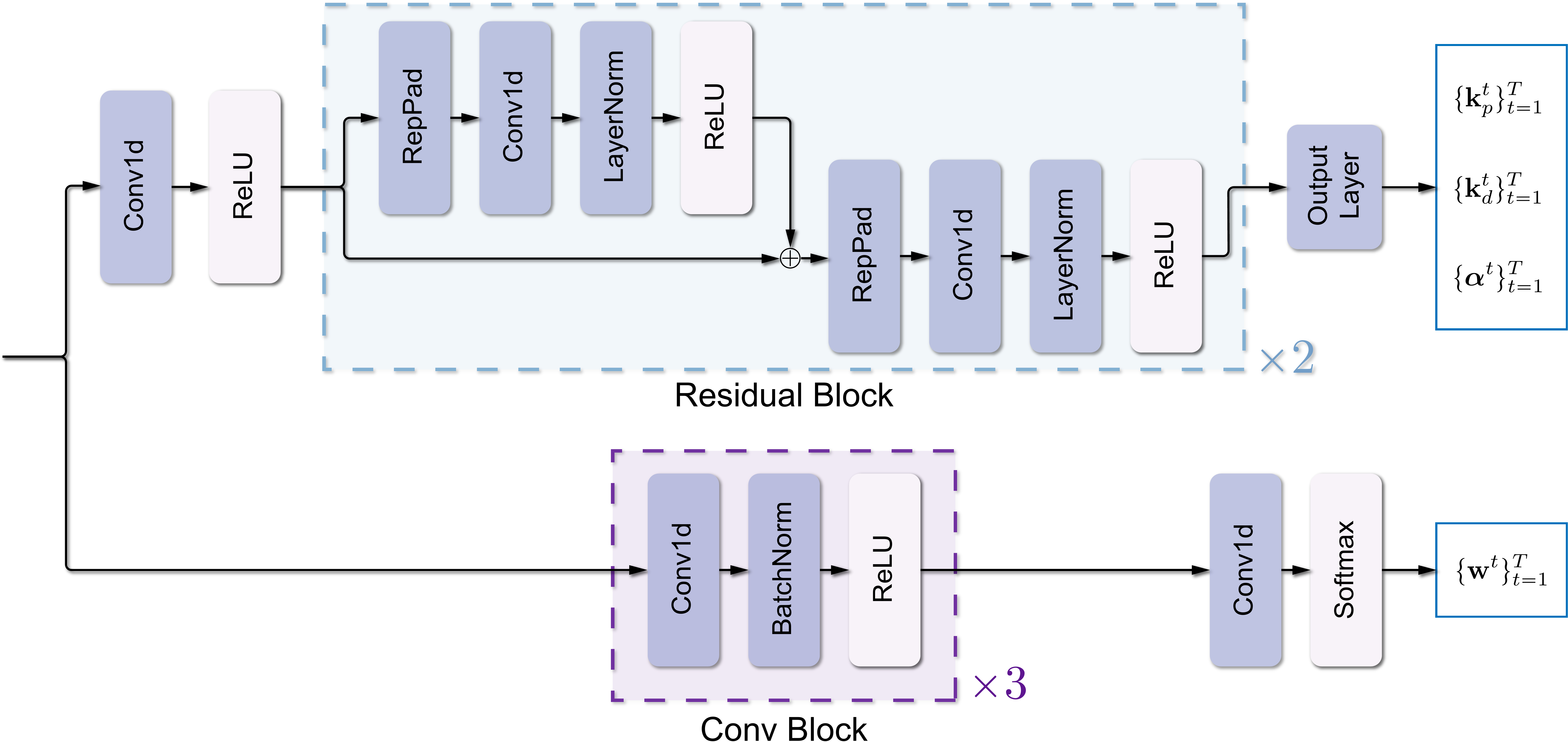}
    \end{center}
    \caption{\textbf{The detailed architecture of attentive PD controller.}}
    \label{fig:pd}
\end{figure}

The detailed network architecture of the attentive PD controller is illustrated in Fig.~\ref{fig:pd}. The kernel size and channel number are set to $3$ and $256$, respectively. To predict the attention weights that satisfy $\sum_{j=1}^T w^{tj} = 1$, we use a softmax layer for normalization.

\section{Details for Contact Annotations}
\label{sec:contact}

To provide supervision signals for contact states, we generate contact annotations on the AMASS dataset~\cite{mahmood2019amass}. Given a video sequence, we first retrieve the toe joints of the body in each frame and then fit the ground plane using the toe positions. Finally, the joints within $4$cm of the ground are labeled as in contact. Following previous works~\cite{shimada2020physcap,shimada2021neural}, we use $N_c = 4$ contact joints, i.e., toes and heels. In daily scenes, the human body might contact the environment with other joints, such as hips and hands. The current model will compensate these undefined contact forces with the residual force, and it is easy for our model to extend to more contact joints.

\section{Comparason with Baselines.}
\label{sec:exp}

To further assess the effectiveness of D\&D, we conduct experiments to compare D\&D with two baselines: the acceleration network (AccNet) and the velocity network (VelNet). Instead of predicting physical properties like forces and contacts, AccNet and VelNet directly predict the pose acceleration and velocity, respectively. The acceleration and velocity are used to control the human motion. The architectures of AccNet and VelNet are similar to DyNet, with an output layer that predicts the pose acceleration and velocity. Other training settings are the same as the proposed D\&D. As shown in Tab.~\ref{table:baseline}, D\&D obtains superior performance to AccNet and VelNet. It demonstrates that the improvement of D\&D comes from the explicit modeling of the physical properties, not controlling human motion with acceleration or velocity.

\begin{table}[h]
    \begin{center}
        \vspace{3mm}
        \caption{\textbf{Comparason with baselines on 3DPW and Human3.6M datasets.}}
        \label{table:baseline}
        \resizebox{\linewidth}{!}
        {
            \begin{tabular}{l|ccc|ccc}
                \toprule
                ~ & \multicolumn{3}{c}{3DPW} & \multicolumn{3}{c}{Human3.6M} \\
                \midrule
                ~ & ~MPJPE~$\downarrow$ & ~PA-MPJPE~$\downarrow$~ & ~ACCEL~$\downarrow$~ & ~MPJPE~$\downarrow$~ & ~PA-MPJPE~$\downarrow$~ & ~ACCEL~$\downarrow$~ \\
                \midrule
                AccNet\, & 101.4 & 72.1 & 10.1 & 90.1 & 58.2 & 14.1 \\
                VelNet\, & 79.3 & 47.0 & 8.4 & 74.1 & 48.8 & 8.5 \\
                D\&D (Ours)\, & \textbf{73.7} & \textbf{42.7} & \textbf{7.0} & \textbf{52.5} & \textbf{35.5} & \textbf{6.1} \\
                \bottomrule
            \end{tabular}
        }
    \end{center}
\end{table}

\section{Physical-based Results on the 3DPW Dataset.}
\label{sec:exp-phys}

We report two physical-based metrics, foot sliding (FS) and ground penetration (GP), to measure the physical plausibility on the 3DPW dataset. We remove the video sequences that contain stairs since it is inaccurate to measure the ground penetration in these cases. Quantitative results are provided in Tab.~\ref{table:phys}. It shows that D\&D can predict physically plausible motion in daily scenes with dynamic camera movements.

\begin{table}[h]
    \begin{center}
        \vspace{3mm}
        \caption{\textbf{Physical-based Results on the 3DPW dataset.}}
        \label{table:phys}
        \begin{tabular}{l|cc}

        \toprule
        Method & ~FS~$\downarrow$~ & ~GP~$\downarrow$~ \\
        \midrule
        HybrIK~\cite{li2021hybrik} & 37.3 & 29.9 \\
        Ours & \textbf{9.8} & \textbf{1.5} \\

        \bottomrule
        \end{tabular}
    \end{center}
\end{table}

\section{Global Trajectory Results on the Human3.6M Dataset.}
\label{sec:exp-accum}

We compare the predicted global trajectory using the simulated moving camera with the results using the static camera on the Human3.6M dataset. Quantitative results are reported in Tab.~\ref{table:gmpjpe}. When testing on the static camera, we directly use the predicted trajectory from HybrIK~\cite{li2021hybrik} in camera coordinates as the global trajectory. We then follow the standard evaluations for open-loop reconstruction (e.g., SLAM and GLAMR~\cite{yuan2022glamr}) to remove the effect of the accumulative error. G-MPJPE and G-PVE are computed using a sliding window (10 seconds) and align the root translation with the GT at the start of each window. We can see that using static camera obtains better results than the closed-loop results of D\&D with the moving camera. When we follow the open-loop protocol to eliminate the accumulative errors, D\&D obtains better results than using the static camera.


\begin{table}[t]
    \begin{center}
        \caption{\textbf{Results of the global trajectory on the Human3.6M dataset.}}
        \label{table:gmpjpe}
        \begin{tabular}{l|cc}

        \toprule
        Method & ~G-MPJPE~$\downarrow$~ & ~G-PVE~$\downarrow$~ \\
        \midrule
        Static Camera & 674.7 & 681.5 \\
        Ours & 785.1 & 793.3 \\
        Ours (open-loop)~ & 525.3 & 533.9 \\

        \bottomrule
        \end{tabular}
    \end{center}
\end{table}

\section{Computation Complexity.}
\label{sec:exp-time}

The whole system is run online by using a sliding window with a length of 16 frames and a stride of 16 frames. The system takes 1349ms for each window (84.3ms for each frame).

\section{Pseudocode}
\label{sec:pseudocode}

The pseudocode of the proposed D\&D is given in Alg.~\ref{alg:dnd}.

\begin{algorithm}[H]
    \caption{\small Pseudocode of D\&D in a PyTorch-like style.
        }
        \label{alg:dnd}
        \definecolor{codeblue}{rgb}{0.25,0.5,0.5}
        \lstset{
            backgroundcolor=\color{white},
            basicstyle=\fontsize{7.2pt}{7.2pt}\ttfamily\selectfont,
            columns=fullflexible,
            breaklines=true,
            captionpos=b,
            commentstyle=\fontsize{7.2pt}{7.2pt}\color{codeblue},
            keywordstyle=\fontsize{7.2pt}{7.2pt},
            escapechar=\&
        }
\begin{lstlisting}[language=python]
# inp_video:    [B, T, 3, H, W]
# init_motion:  [B, T, 75]
# betas:        [B, 10]
init_motion, betas = kinematics_net(inp_video)

# Dynamics Networks 
eta, a_ine, w_ine, g, lambda, p = DyNet(init_motion)
kp, kd, alpha, w = AttenPDController(init_motion)

# Initialize pose at time 0
final_q, final_dq = [q_0], [dq_0]
final_qtrans, final_dqtrans = [qtrans_0], [dqtrans_0]
rot_cam = 0

# Analytical Computation
for t in range(time_len - 1):
    current_q, current_dq = final_q[t], final_dq[t]
    current_qtrans, current_dqtrans = final_qtrans[t], final_dqtrans[t]

    # Compute Inertia Matrix and Jacobian
    M, Jv, Jw, m = get_jacobian(current_q, betas)
    # Compute physical torques
    target_q_state = torch.sum(w[t + 1] * initial_motion, dim=1)
    tau = kp[t] * (target_q_state - current_q) - kd[t] * current_dq + alpha[t]

    I = compute_inertia(m, Jv[t], a_ine[t], w_int[t])
    h_g = compute_gravity(m, Jv[t], g[t])
    h_grf = PCT(Jv, p[t], lambda[t])

    # Inertial Forward Dynamics
    ddq = M.inv().matmul(tau + h_g + h_grf + I)

    rot_cam = rot_cam + w_ine * delta_t

    # Trajectory Forward Dynamics
    ddqtrans = rot_cam.to_matrix().T.matmul(eta[t] + h_grf[:, 0:3] + h_g[:, 0:3]) / m[:, 0]

    # Constrained Update
    dq = current_dq + ddq * delta_t
    dqtrans = current_dqtrans + ddqtrans * delta_t

    dq, dqtrans = cvx_layer(dq, dqtrans, Jv[t], p[t], rot_cam)

    q = current_q + dq * delta_t
    qtrans = current_qtrans + dqtrans * delta_t

    final_q.append(q)
    final_dq.append(dq)
    final_qtrans.append(qtrans)
    final_dqtrans.append(dqtrans)

final_q = torch.cat(final_q, dim=1)
final_qtrans = torch.cat(final_qtrans, dim=1)

\end{lstlisting}
\end{algorithm}

\section{Qualitative Results}
\label{sec:res}

\begin{figure}[ht]
    \begin{center}
    \includegraphics[width=.98\linewidth]{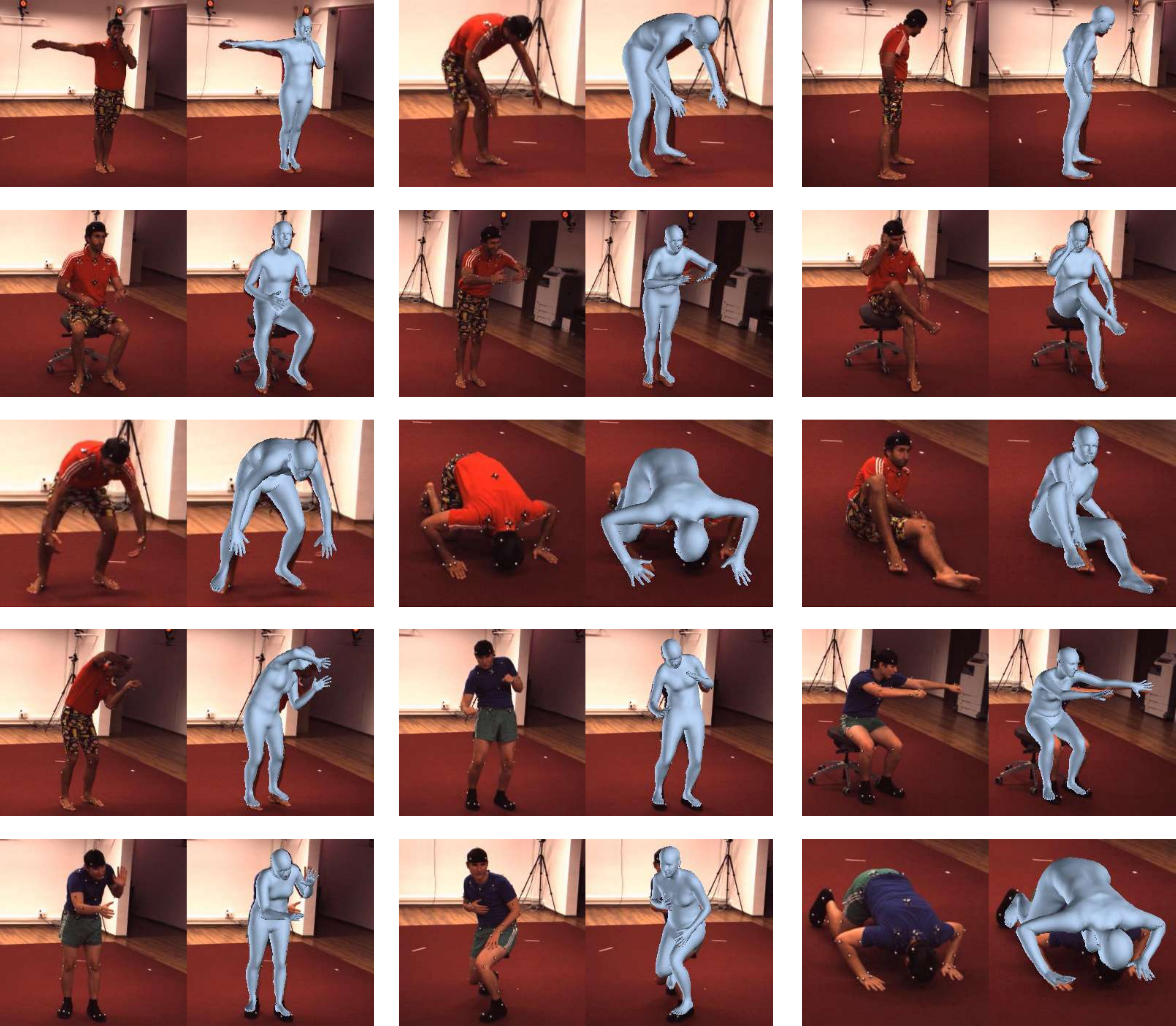}
    \end{center}
    \caption{\textbf{Qualitative results on the Human3.6M dataset.}}
    \label{fig:qual-h36m}
\end{figure}

\begin{figure}[ht]
    \begin{center}
    \includegraphics[width=.98\linewidth]{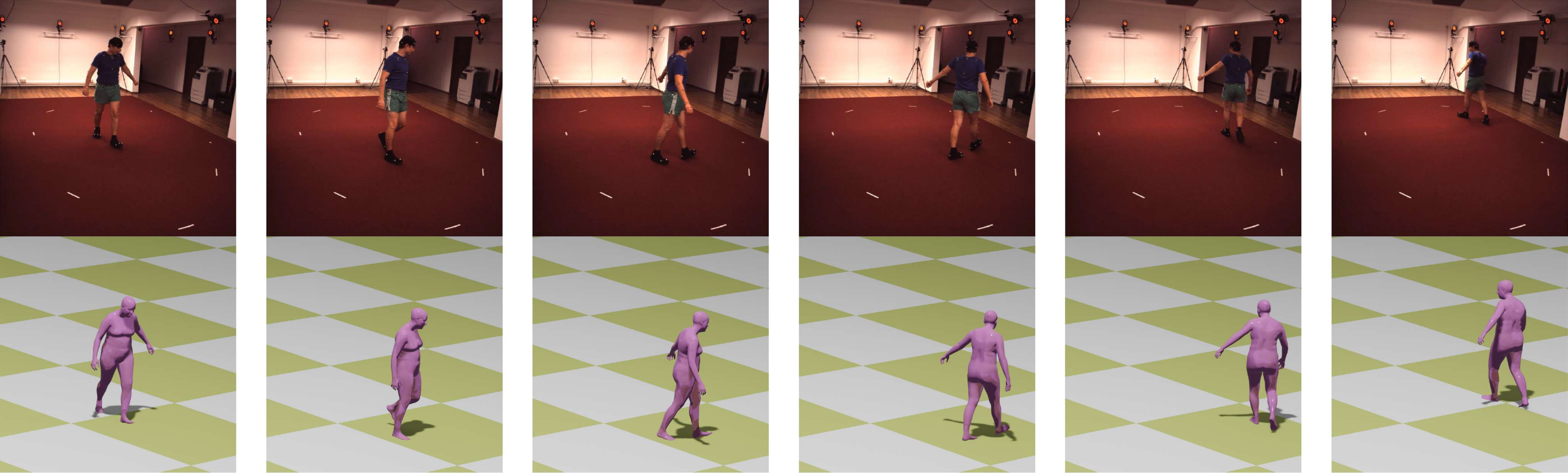}
    \end{center}
    \caption{\textbf{Qualitative results on the Human3.6M dataset.}}
    \label{fig:qual-h36m-blender}
\end{figure}

\begin{figure}[ht]
    \begin{center}
    \includegraphics[width=.98\linewidth]{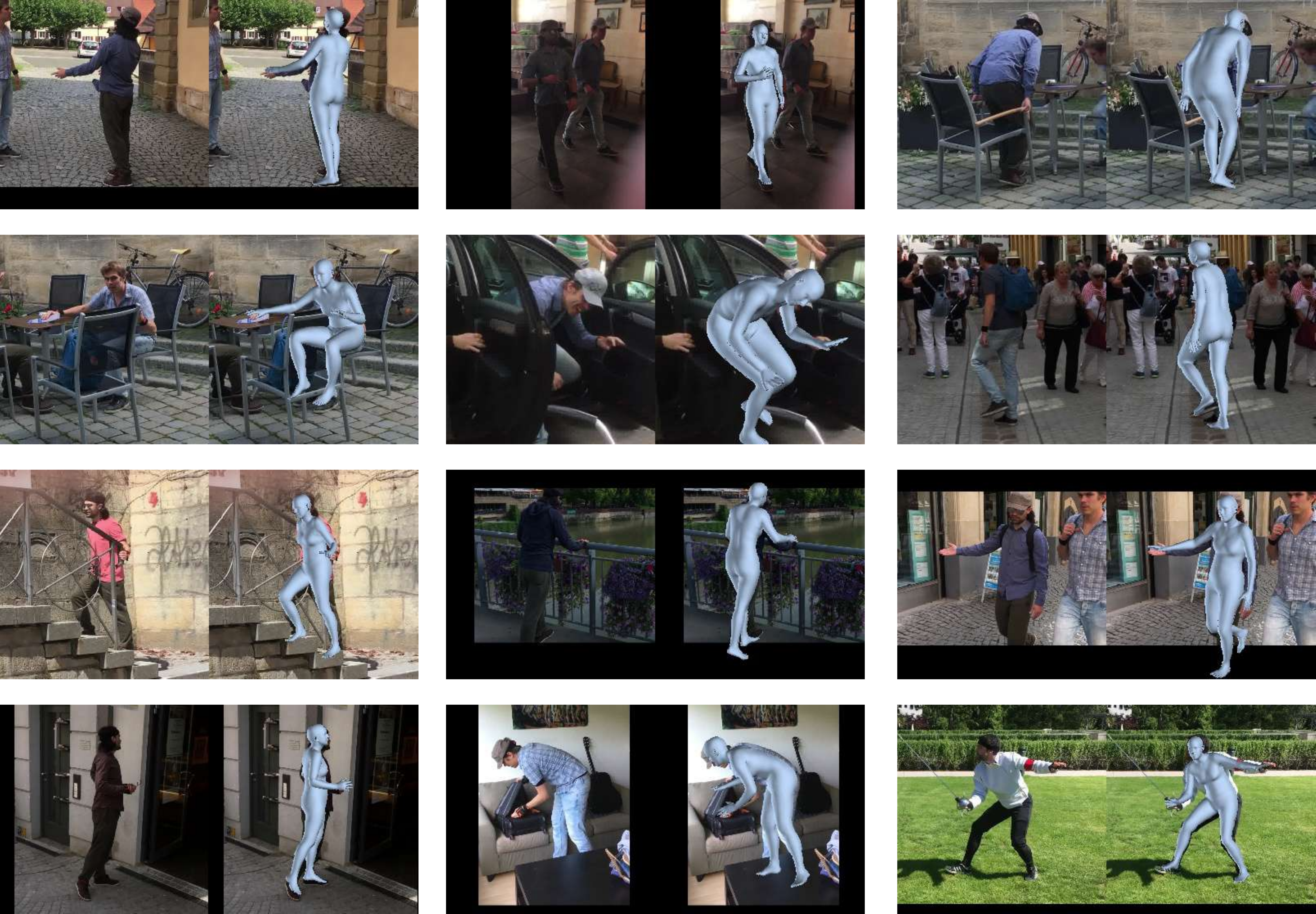}
    \end{center}
    \caption{\textbf{Qualitative results on the 3DPW dataset.}}
    \label{fig:qual-3dpw}
\end{figure}

\begin{figure}[ht]
    \begin{center}
    \includegraphics[width=.98\linewidth]{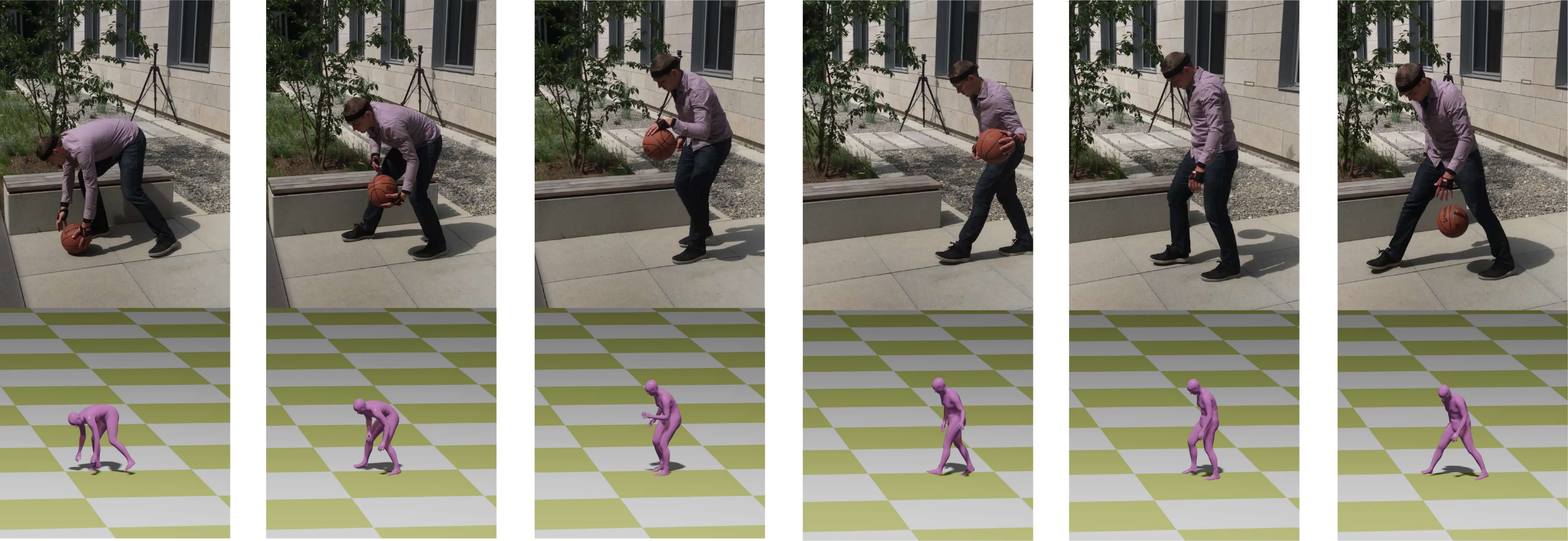}
    \end{center}
    \caption{\textbf{Qualitative results on the 3DPW dataset.}}
    \label{fig:qual-3dpw-blender1}
\end{figure}

\begin{figure}[ht]
    \begin{center}
    \includegraphics[width=.98\linewidth]{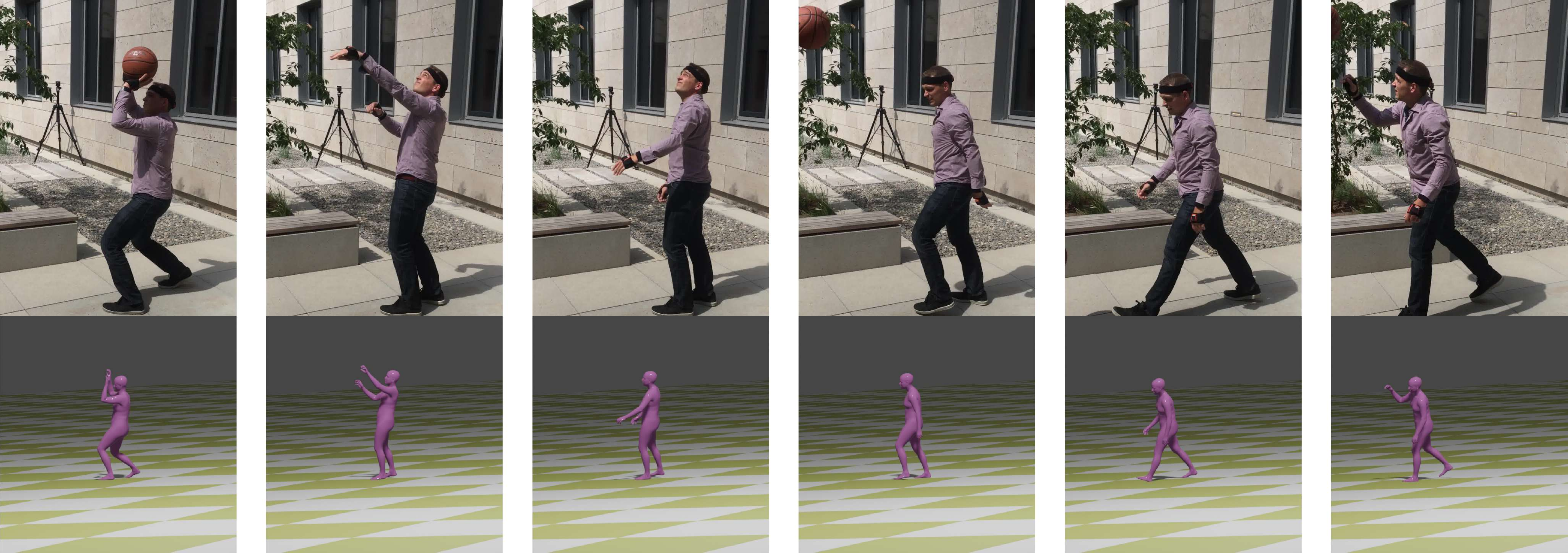}
    \end{center}
    \caption{\textbf{Qualitative results on the 3DPW dataset.}}
    \label{fig:qual-3dpw-blender2}
\end{figure}

\begin{figure}[ht]
    \begin{center}
    \includegraphics[width=.98\linewidth]{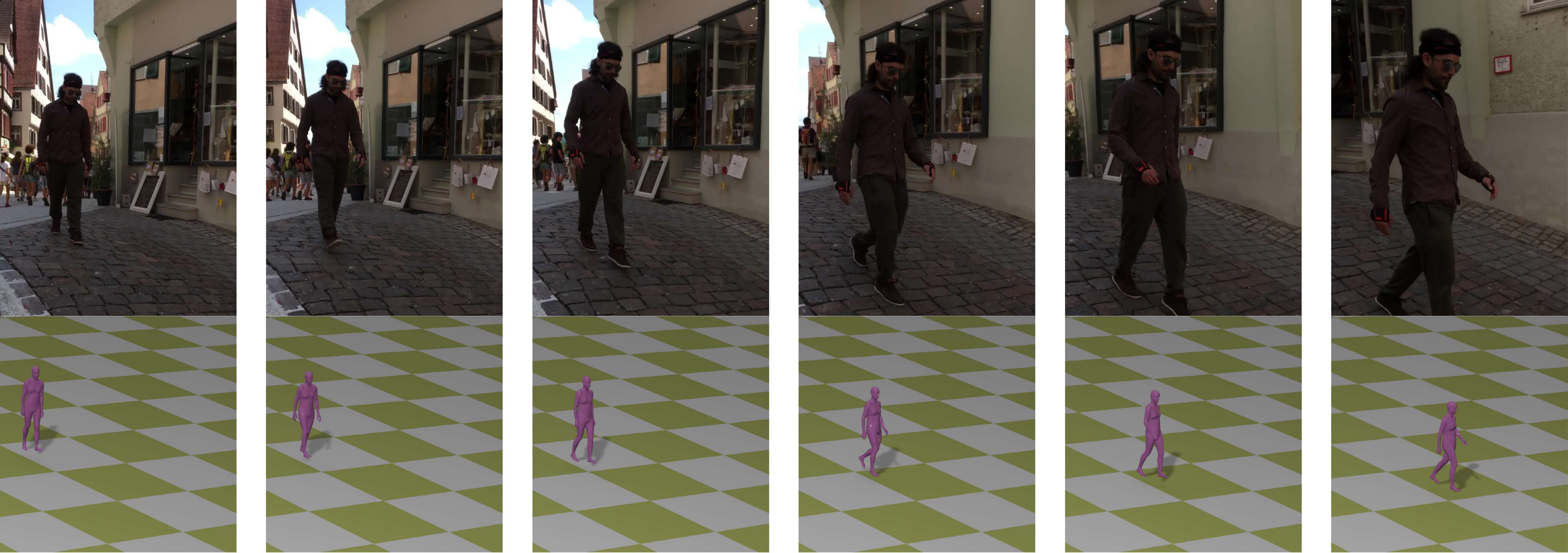}
    \end{center}
    \caption{\textbf{Qualitative results on the 3DPW dataset.}}
    \label{fig:qual-3dpw-blender3}
\end{figure}

\begin{figure}[ht]
    \begin{center}
    \includegraphics[width=.98\linewidth]{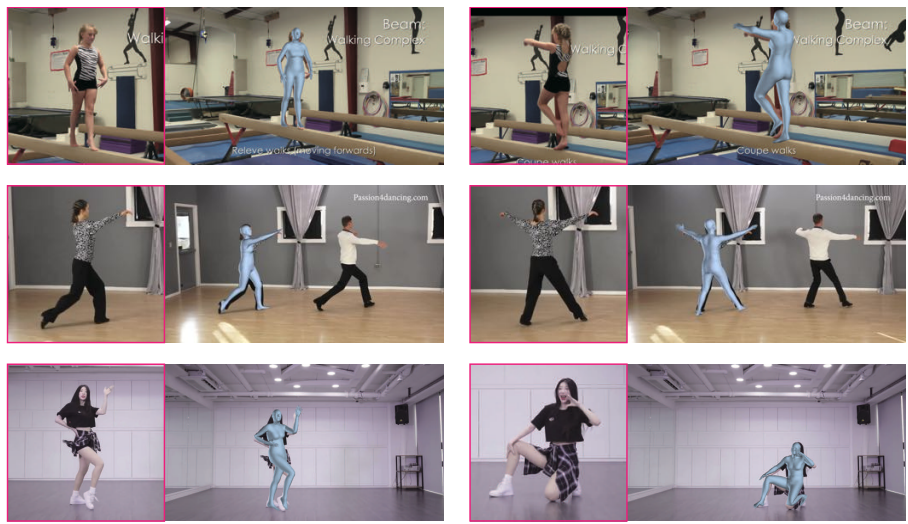}
    \end{center}
    \caption{\textbf{Qualitative results on the Internet videos.}}
    \label{fig:internet}
\end{figure}

Additional qualitative results are shown in Fig.~\ref{fig:qual-h36m}, Fig.~\ref{fig:qual-h36m-blender} (the Human3.6M dataset), Fig.~\ref{fig:qual-3dpw}, Fig.~\ref{fig:qual-3dpw-blender1}, Fig.~\ref{fig:qual-3dpw-blender2}, Fig.~\ref{fig:qual-3dpw-blender3} (the 3DPW dataset), and Fig.~\ref{fig:internet} (the Internet videos).

\end{document}